\documentclass{bmcart}

\usepackage{amsthm,amsmath}
\usepackage{adjustbox}
\usepackage{graphicx}
\RequirePackage[numbers]{natbib}
\usepackage{algorithm}
\usepackage[noend]{algpseudocode}

\usepackage{amssymb}

\startlocaldefs
\endlocaldefs

\begin{document}

\begin{frontmatter}

\begin{fmbox}


\title{Exploiting Prunability for Person Re-Identification}

\author[
  addressref={aff1},
]{\inits{H.M}\fnm{Hugo} \snm{Masson}}
\author[
  addressref={aff1,aff2},                   
  corref={aff1},                       
  email={amran.apece@gmail.com}   
]{\inits{A.B.}\fnm{Amran} \snm{Bhuiyan}}
\author[
  addressref={aff1},
]{\inits{L.T}\fnm{Le Thanh} \snm{Nguyen-Meidine}}
\author[
  addressref={aff2},
]{\inits{M.J}\fnm{Mehrsan} \snm{Javan}}
\author[
  addressref={aff2},
]{\inits{P.S}\fnm{Parthipan} \snm{Siva}}
\author[
  addressref={aff1},
]{\inits{I.A}\fnm{Ismail Ben } \snm{Ayed}}
\author[
  addressref={aff1},
]{\inits{E.G}\fnm{Eric} \snm{Granger}}

\address[id=aff1]{
  \orgdiv{LIVIA, Department of Systems Engineering},             
  \orgname{\'Ecole de technologie sup\'erieure},          
  \city{Montreal},                              
  \cny{Canada}\\                                    
  \orgname{*First three authors contributed equally to this paper}
}
\address[id=aff2]{%
  \orgdiv{SLIQ Labs},
  \orgname{Sportlogiq Inc.},
  \city{Montreal},
  \cny{Canada}
}

\end{fmbox}

\begin{abstractbox}
\begin{abstract} 
Recent years have witnessed a substantial increase in the deep learning (DL) architectures proposed for visual recognition tasks like person re-identification, where individuals must be recognized over multiple distributed cameras. Although these  architectures have greatly improved the state-of-the-art accuracy, the computational complexity of the CNNs commonly used for feature extraction remains an issue, hindering their deployment on platforms with limited resources, or in applications with real-time constraints. There is an obvious advantage to accelerating and compressing DL models without significantly decreasing their accuracy. However, the source (pruning) domain differs from operational (target) domains, and the domain shift between image data captured with different non-overlapping camera viewpoints leads to lower recognition accuracy. In this paper, we investigate the prunability of these architectures under different design scenarios. This paper first revisits pruning techniques that are suitable for reducing the computational complexity of deep CNN networks applied to person re-identification. Then, these techniques are analysed according to their pruning criteria and strategy, and according to different scenarios for exploiting pruning methods to fine-tuning networks to target domains. Experimental results obtained using DL models with ResNet feature extractors, and multiple benchmarks re-identification datasets, indicate that pruning can considerably reduce network complexity while maintaining a high level of accuracy. In scenarios where pruning is performed with large pre-training or fine-tuning datasets, the number of FLOPS required by  ResNet architectures is reduced by half, while maintaining a comparable rank-1 accuracy (within 1\% of the original model). Pruning while training a larger CNNs can also provide a significantly better performance than fine-tuning smaller ones.
\end{abstract}



\end{abstractbox}
%

\end{frontmatter}

\section{Introduction}
\label{S:1}

Deep learning (DL) architectures like the convolutional neural network (CNN) have achieved state-of-the-art accuracy across a wide range of visual recognition tasks, at the expense of growing complexity (deeper and wider networks) that require more training samples and computational resources. In order to deploy these architectures on compact platforms with limited resources (e.g. embedded systems, mobile phones, portable devices), and for real-time processing (e.g. video surveillance and monitoring, virtual reality), their time and memory complexity and energy consumption should be reduced~\citep{huang2017speed}. Consequently, there is a growing interest in effective methods able to accelerate and compress deep networks. 

\begin{figure}[t!]
\centering\includegraphics[width=1\linewidth,trim={3.5in 0 0 0},clip]{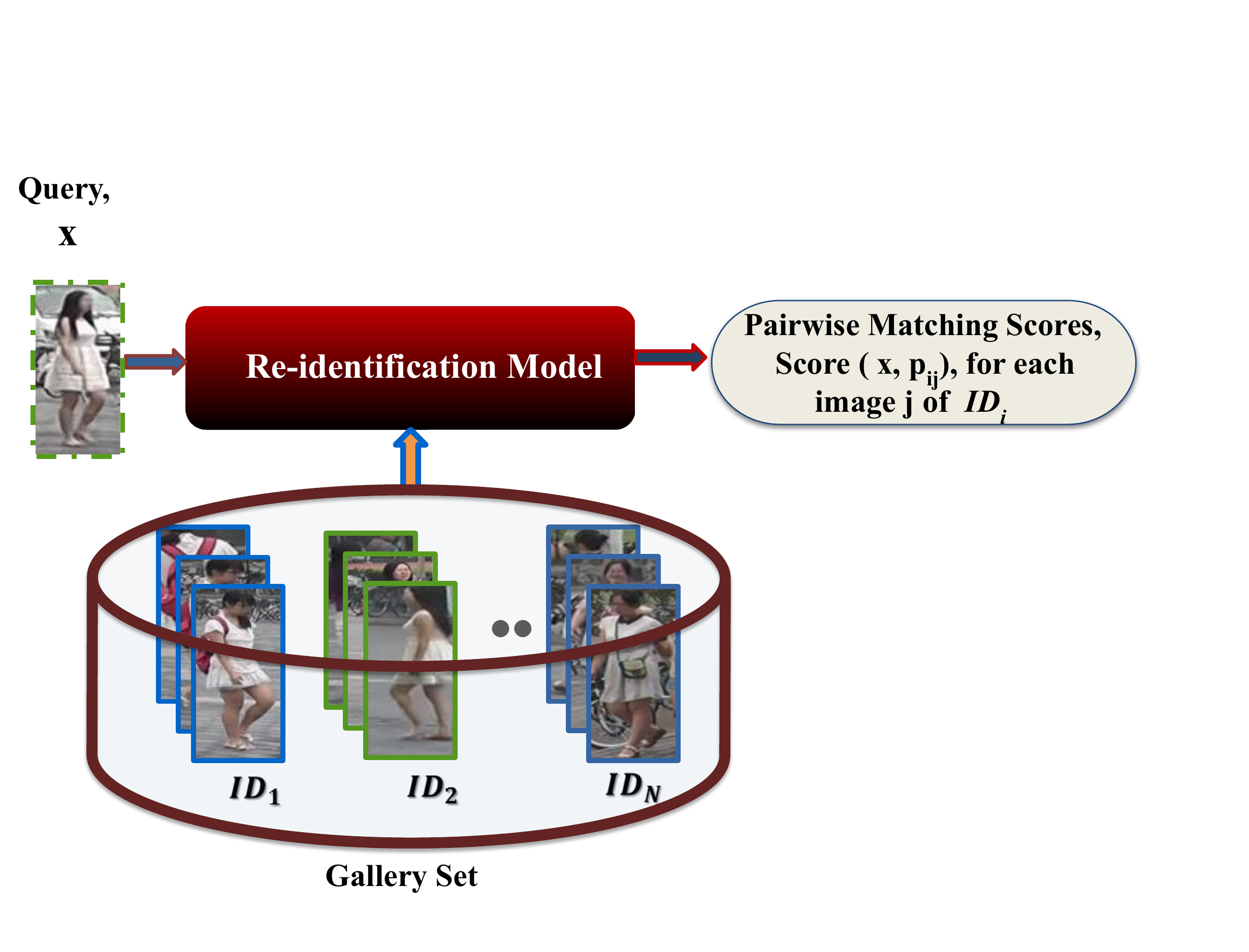}
\caption{Illustration of a typical person re-identification system.}
\label{Person_ReID}
\end{figure}

Providing a reasonable trade-off between accuracy and efficiency has become an important concern in person re-identification (ReID), a key function needed in a wide range of video analytics and surveillance applications. Systems for person ReID typically seek to recognize the same individuals that previously appeared over a non-overlapping network of video surveillance cameras (see illustration in Fig.~\ref{Person_ReID}). These systems face many challenges in real-world application that are either related to the image data or the network architecture. Data related challenges that affect the ReID accuracy include the limited availability of annotated training data, ambiguous annotations, domain shifts across camera viewpoints, limitations of person detection and tracking techniques, occlusions, variations in pose, scale and illumination, and low-resolution images.

To address these issues, state-of-the-art DL models (e.g., deep Siamese networks) for person ReID often rely CNNs for feature extraction to learn an embedding in end-to-end fashion, where similar image pairs (with the same identity) are close to each other, while dissimilar image pairs (with different identities) are distant from each other~\citep{ahmed2015improved,hermans2017defense,varior2016gated,chen2017beyond,geng2016deep,cheng2016person,liu2017end,bhuiyan2020pose,bhuiyan2014person}. While state-of-the-art approaches can provide a high level of accuracy, achieving this performance comes with the cost of millions or even billions of parameters, a challenging training procedure, and the requirement for GPU acceleration.  For instance, the ResNet50 CNN~\citep{he2016deep}, with it 50 convolutional layers, contains about 23.5M parameters (stored in 85.94MB of memory), and requires 6.3 billion floating point operations (FLOPs) to process a color image of size $256 \times 128 \times 3$. The complexity of these networks limit their deployment in many real-time applications, or on resource-limited platforms. Consequently, there has been a great deal of interest by  the computer vision and machine learning communities to develop methods able to accelerate and compress such networks, as well as other DL architectures, without compromising their predictive accuracy. 

The time complexity of a CNN generally depends more on the convolutional layers, while the fully connected layers contain the most of the parameters (memory complexity). Therefore, the CNN acceleration methods typically target lowering the complexity of the convolutional layers, while the compression methods usually target reduced complexity of the fully connected layers~\citep{han2015deep, han2015learning}.  State-of-the-art approaches for acceleration and compression of deep neural networks can be divided into five categories -- low-rank factorization, transferred convolutional channels, knowledge distillation, quantization and pruning.

Low-rank factorization approaches~\citep{lu2017fully, rigamonti2013learning,denton2014exploiting,jaderberg2014speeding,lebedev2014speeding,tai2015convolutional} accelerate CNNs by performing matrix decomposition to estimate information parameters of a network. However, low-rank approaches suffer from a number of issues -- computationally expensive matrix decomposition, layer by layer low-rank approximation that diminishes the possibility of global compression, and extensive model retraining to achieve convergence. 
Some network acceleration and compression  approaches~\citep{cohen2016group,van2016deep,shang2016understanding,dieleman2016exploiting}, categorized as transferred convolutional channels, design special structural convolutional channels to reduce the parameter space, which eventually improves computational efficiency, but  transfer assumptions are sometimes too strong that makes the learning process unstable.

Knowledge distillation approaches~\citep{bucilua2006model,hinton2015distilling,luo2016face,chen2015net2net} train a smaller or shallow deep network (the student) using distilled knowledge of a larger deep network, called teacher. These approaches can yield improvements in term of sparsity and generalization of the student networks, but can  only  be  applied  to  classification  tasks and the bounded assumption of this approach leads to inferior performance while compares to other types of approaches. 
A deep neural network can be accelerated by reducing the precision of its parameters. Using  quantization approaches, each parameter of a network is represented with a reduced bit rate, either by reducing the precision, employing a lookup table, or combining similar values. Most of the quantization approaches~\citep{han2015deep,gong2014compressing,chen2015compressing} require extra computational time to access a look up table, or for decoding such that the original value is restored.  
In contrast, pruning seeks to reduce the number of connections or retrain either the whole, or part, of the network with a freshly trained replacement. Pruning methods typically focus on selecting and removing the weights or channels with the least impact on performance. Thus, in addition to accelerating and compressing the network, pruning methods can provide the additional benefits such as addressing the overfitting problem and thus improve generalization. Therefore, pruning approaches have drawn a great deal of attention from the network compression community. Challenges of pruning include the lack of data for pruning during the fine-tuning phase, the computational complexity associated with retraining after a pruning phase, and the reduction of capacity to learn of a model, which can impact the accuracy when the learning step is done on the pruned model.

This paper focuses on pruning techniques~~\citep{lecun1990optimal, hassibi1993second, han2015learning, Molchanov, HaoLi} since they are among the most widely used for acceleration and compression of deep neural networks, and have been shown their effectiveness on well-known CNNs, and for several general image classification problems like CIFAR10, MNIST and ImageNet. State-of-the-art pruning techniques can be categorized according to their pruning criteria to select channels, and to their strategy to reduce channels, and are suitable for compressing DL models like Siamese networks for applications in person Re-ID. In particular,  state-of-the-art techniques can be categorized using criteria based on weights and on feature maps. We also distinguish techniques according to pruning strategy, pruning techniques can also be distinguished among -- those that (1) prune once and then fine-tune, (2) prune iteratively on trained model, (3) prune using regularization, (4) prune by minimizing the reconstruction errors, and (5) prune progressively. 
 
This paper revisits the pruning techniques that are suitable for reducing the computational complexity of CNNs applied to person Re-ID. These techniques are then analysed according to their pruning criteria and strategy, and according to different design scenarios. Different design pipelines or scenarios are proposed to leverage these state-of-the-art pruning methods during pre-training and/or fine-tuning. A typical design scenario consists of four stages: (1) training a CNN with a large-scale dataset from the source domain (i.e., ImageNet), (2) prune the trained large model based on some criterion to select channels to be eliminated, (3) re-train the pruned network to regain the accuracy, and finally, (4) fine-tune the re-trained network using a limited dataset from the target application. A common assumption with this design scenario is that training a large and over-parameterized CNNs, using a large-scale dataset, is necessary to provide a discriminant feature representation. The pruning process used to select and reduce the network will yield a set of redundant channels that does not significantly reduce accuracy. Under this scenario, a CNNs for Re-ID would therefore over-train on a smaller network from scratch~\citep{HaoLi, Entropy_Pruning, ChannelPruning,Pro_Soft_Pruning}. Thus, most of the approaches in literature tend to prune channels of a fine-tuned network, rather than a pre-trained network. This paper present other design scenarios that apply when pruning networks that have been pre-trained on large dataset, and that require a fine-tuning to a given target domain. 

Finally, this paper presents an extensive experimental comparison of different pruning techniques and relevant design scenarios on three benchmark person reID datasets -- the Market-1501 ~\citep{market1501}, CUHK03-NP~\citep{cuhk03}, and DukeMTMC-reID~\citep{DukeMTMC}. Pruning techniques are compared in terms of accuracy and complexity on different DL architectures with ResNet feature extractors, with different reID applications in mind.

The rest of the paper is organized as follows. Section~\ref{S:3} provides some background on DL models for person ReID. Section~\ref{S:2} provides a survey of the state-of-the-art techniques for pruning CNNs. Finally, Sections~\ref{S:4} and~\ref{S:5} described the experimental methodology (benchmark datasets, protocol and performance measures), and comparative results, respectively.

\section{Deep Neural Networks for Person Re-Identification}
\label{S:3}

State-of-the-art techniques for person ReID mostly rely on two types of losses: metric learning loss and multi-class classification loss. With the first type, a dataset with images from different individuals is learned using a Siamese network that optimizes a metric loss function (such as contrastive loss, triplet loss, quadruplet loss, hard-aware point-to-set (HAP2S) loss)~\citep{yi2014deep,ahmed2015improved,varior2016gated, cheng2016person,TRIPLET_REID} to provide a feature embedding for pair-wise similarity matching. With the second type, ReID approaches based on multi-class classification loss (such as softmax or cross-entropy loss)~\citep{Past_Present_Future,sun2018beyond,su2017pose,yao2019deep,zhao2017deeply} learn part-based local features to form more informative feature descriptor, also known as ID-loss in ReID community. 

\begin{table*}[b!]
\vspace{-0.1cm}
    \caption{Common loss functions applied in person ReID.}
   \hspace{+0.05cm}
   \centering
 \begin{adjustbox}{width=1\textwidth} 
  \begin{tabular}{|l|l|}
    \hline
    \bf{Category}        & \bf{Loss Function}        \\ \hline \hline
    
&  Contrastive Loss ~\citep{varior2016gated}: \ 	
   \[ \begin{aligned}
	\mathcal{L}_{\mbox{CE}}= \frac{1}{2N} \sum_{i=1}^N \big[ (1- y_i) d^2 \left(\textbf{f}_{i,1},\textbf{f}_{2,i}\right) + (y_i)\{\max(0, m- d^2 \left(\textbf{f}_{1,i},\textbf{f}_{2,i}\right))\} \big]
	\end{aligned}\] \\  \cline{2-2} 

&  Triplet ~\citep{ding2015deep}: \ 
	\[\begin{aligned}
	\mathcal{L}_{\mbox{T}}= \frac{1}{N_T} \sum_{\substack{\text{a,p,n} \\ \text{$y_a$=$y_p$$\neq$$y_n$}}} \big[d \left(\textbf{f}_a,\textbf{f}_p\right) -d \left(\textbf{f}_a,\textbf{f}_n\right)\big]_{+}
	\end{aligned} \]\\ \cline{2-2}
	
&  Triplet Loss with Margin ~\citep{cheng2016person}: \
	\[\begin{aligned}
	\mathcal{L}_{\mbox{T}}= \frac{1}{N_T} \sum_{\substack{\text{a,p,n} \\ \text{$y_a$=$y_p$$\neq$$y_n$}}} \big[m+d \left(\textbf{f}_a,\textbf{f}_p\right) -d \left(\textbf{f}_a,\textbf{f}_n\right)\big]_{+}
	\end{aligned} \]\\  \cline{2-2}
	
Metric Learning  &  Semi-Hard Triplet ~\citep{TRIPLET_REID}: \
	\[\begin{aligned}
	\mathcal{L}_{\mbox{TBH}}&= \frac{1}{N_s} \sum_{a=1}^{N_s}\big[m+ \operatorname*{max}_{y_p =y_a} d \left(\textbf{f}_a,\textbf{f}_p\right)- \operatorname*{min}_{y_p \neq y_a} d \left(\textbf{f}_a,\textbf{f}_n\right)]_{+}
	\end{aligned} \]\\  \cline{2-2}
	
&  Quadruplet ~\citep{chen2017beyond}: \ 
	\[\begin{aligned}
	\mathcal{L}_{\mbox{quad}}= \frac{1}{N} \sum_{\substack{\text{a,p,n} \\ \text{$y_a$=$y_p$$\neq$$y_n$}}} \big[m_1+d \left(\textbf{f}_a,\textbf{f}_p\right) -d \left(\textbf{f}_a,\textbf{f}_n\right)\big]_{+} \\
	\ \ \  +
	\frac{1}{N} \sum_{\substack{\text{a,p,n,k} \\ \text{$y_a$=$y_p$$\neq$$y_n$$\neq$$y_k$}}} \big[m_2+d \left(\textbf{f}_a,\textbf{f}_p\right) -d \left(\textbf{f}_n,\textbf{f}_k\right)\big]_{+} 
	\end{aligned} \]\\ \cline{2-2}

&  HAP2S ~\citep{yu2018hard}: \
	\[\begin{aligned}
	\mathcal{L}_{\mbox{HAP2S}}&= \frac{1}{N_s} \sum_{a=1}^{N_s}\big[m+ \operatorname*{max}_{y_p =y_a} d \left(\textbf{f}_a,\textbf{S}_p\right)- \operatorname*{min}_{y_p \neq y_a} d \left(\textbf{f}_a,\textbf{S}_n\right)]_{+}
	\end{aligned} \]\\ \cline{2-2}

&  Magnet ~\citep{wojke2018deep}: \
	\[\begin{aligned}
	\mathcal{L}_{\mbox{mag}}= - \frac{1}{N} \sum_{i=1}^{N} \big[ \log \frac{e^{-\frac{1}{2{\sigma}^2} d\left(\textbf{f}_i, {\mu}(\textbf{f}_i)\right)-m}}{ \sum_{k=1}^{C} e^{-\frac{1}{2{\sigma}^2} d\left(\textbf{f}_i, {\mu}_i^k
	\right)}} \big]_{+}
	\end{aligned}	\]\\ \hline \hline

&  Cross-Entropy ~\citep{ahmed2015improved,wu2016personnet,cuhk03}: \
	\[\begin{aligned}
	\mathcal{L}_{\mbox{CE}}= - \frac{1}{N} \sum_{i=1}^{N} \log \frac{e^{\textbf{W}_{y_i}^{T} \textbf{f}_i}}{ \sum_{k=1}^{C} e^{\textbf{W}_k^{T} \textbf{f}_i}}
	\end{aligned} \]\\ \cline{2-2}
Classification &  Cosine Softmax ~\citep{wojke2018deep}: \
	\[\begin{aligned}
	\mathcal{L}_{\mbox{CCE}}= - \frac{1}{N} \sum_{i=1}^{N} \log \frac{e^{\kappa . \Tilde{\textbf{W}_{y_i}^{T}} \Tilde{\textbf{f}_i}}}{ \sum_{k=1}^{C} e^{\kappa . \Tilde{\textbf{W}_k^{T}} \Tilde{\textbf{f}_i}}}
	\end{aligned}\] \\ \cline{2-2}
&  Part-Based Cross-Entropy ~\citep{sun2018beyond,su2017pose,yao2019deep,zhao2017deeply}: \ 
	\[\begin{aligned}
	\mathcal{L}_{\mbox{PCE}}=  \sum_{p=1}^{P} \mathcal{L}_{CE}^p 
	\end{aligned} \]\\ \hline

  \end{tabular}
  \end{adjustbox}
   \vspace{-0.3cm}
  \label{tab:1}
\end{table*}

Table~\ref{tab:1} provides a summary of common loss functions from both categories applied in person ReID. For all the losses, $d$ represents the Euclidean distance, $m, m_1, m_2$ denotes the margin parameters and  $\big[ . \big] _{+} = \max(.,0)$. In this table, $ \textbf{X} = \{\textbf{x}\}_{i=1}^N$ is a training mini-batch with labels $\{y_i\}_{i=1}^N$. For  contrastive loss, network transforms the pair of input images ${\textbf{x}_{1,i}, \textbf{x}_{2,i}}$ into feature embeddings ${\textbf{f}_{1,i}, \textbf{f}_{2,i}}$. The labels are either $y_i = 0$ for positive pairs, or $y_i = 1$ for negative pairs. For triplet loss, we sample $(\textbf{x}_a, \textbf{x}_p, \textbf{x}_n)$ where the anchor and the positive $\textbf{x}_a$ are two images from the same person, while the negative $\textbf{x}_n$ is an image from another person. The corresponding feature embeddings are $(\textbf{f}_a, \textbf{f}_p, \textbf{f}_n)$. For quadruplet loss, we sample $(\textbf{x}_a, \textbf{x}_p, \textbf{x}_n, \textbf{x}_k)$ where the anchor and positive $\textbf{x}_a$ are two images from the same person, while the negative and $\textbf{x}_k$ are the images from different persons. The corresponding feature embeddings are $(\textbf{f}_a, \textbf{f}_p, \textbf{f}_n,\textbf{f}_k)$. For HAP2S loss,  $\textbf{S}_p$ and $\textbf{S}_n$ denote the set of positive and negative samples respectively with respect to the anchor $\textbf{f}_a$;  for magnet loss, $C$ is the number of classes (individual), $\mu$ is the sample mean of class $y$,  and $ {\sigma}^2$ is the variance of all samples away from their class mean. Belongs to classification loss categoy, in cross-entropy loss, $\textbf{W}_{y_i}$ is the weight vector of the fully connected layer  with feature embedding $\textbf{f}_i$. For cosine softmax loss, $\Tilde{\textbf{W}}_{y_i}$ and $\Tilde{\textbf{f}}_i$ denotes the normalized weight and feature vector respectively; and for part-based cross entropy loss, $L_{CE}^p$ represents the cross entropy loss of individual part, $P$.

\subsection{Metric Loss:}

The idea of using deep Siamese networks for biometric authentication and verification originate from Bromeley et al.~\citep{bromley1994signature}, where two sub-networks with shared weights encode feature embeddings for pairwise matching between a query and reference (gallery) images. These networks were first used in~\citep{yi2014deep} for ReID that employ three feature extraction sub-networks for deep feature learning. Then, various deep learning architectures were proposed to learn discriminative feature embeddings. Most of these architectures~\citep{ding2015deep,varior2016gated,cheng2016person,TRIPLET_REID,chen2017beyond,liu2017end, yu2018hard,wojke2018deep} employ end-to-end training, where both feature embedding and metric are learned as a joint optimization problem.

There are a number of metric learning losses that are widely used for optimizing deep ReID architectures. Contrastive loss is used in \citep{varior2016gated} to optimize a Siamese network that minimizes  the distance  between  samples  of  the  same  class  and  forces  a margin between samples of different classes. Triplet loss in ReID is first used in ~\citep{ding2015deep} that directly optimizes an embedding layer in euclidean space which compares the relative distances of three training samples, namely an anchor image, a positive image sample from the same individual, and a negative sample from a different individual. In~\citep{cheng2016person}, original triplet loss is modified by adding an additional positive-pair constraint. A different version of triplet loss, named as quadruplet loss proposed in~\citep{chen2017beyond}, which enlarges inter-class variations and reduces intra-class variation. Hermans et al.~\citep{TRIPLET_REID} extends the triplet loss by designing a simple semi-hard mining that selects the hardest positive and hardest negative of each anchor in a mini-batch. In~\citep{yu2018hard}, a soft hard-sample mining scheme is proposed  by adaptively assigning weights to hard samples. On the other hand, magnet loss~\citep{wojke2018deep}  is formulated as a negative log-likelihood ratio between the correct class and all other classes, but also forces a margin between samples of different classes. A through study of all the state-of-the-art metric learning losses for ReID suggests that triplet loss is the most widely used loss to optimize the deep ReID architecture. And among all the different version of the triplet losses, the semi-hard mining based triplet loss proposed by Herman et al.\citep{TRIPLET_REID} is the simplest and efficient that does not require to change the backbone architecture to get the final feature embedding.

\subsection{Multi-Class Classification Loss:}

There has been an alternative trend that addresses the ReID problem as multi-class classification problems, where each ID considers as a class. The objective of classification loss is to determine whether each input pair of images are same or not which makes full use of the ReID label with the predicted one from the classification networks. Some of the state-of-the-art classification based ReID approaches~\citep{ahmed2015improved,wu2016personnet,cuhk03} employ cross-entropy loss for image pairs in their network that takes pairwise images as inputs, and output the verification probability. Some other state-of-the-art ReID approaches~\citep{varior2016siamese,wang2016joint} used margin-based loss to keep the largest possible separation between the positive and negative pairs.

Recently, many works focus to learn local part-based feature which adopts the simple classification loss based network performed on multiple local parts of a single image. Most of these approaches take into account the local features either from the human body part or by diving the global features to obtain discriminative part based feature representations. State-of-the-art approaches~\citep{su2017pose,yao2019deep,zhao2017deeply} rely on diving the human body parts based on either external pose estimation or external semantic segmentation that leverage the semantic partitions for deeply learned part based features. However, they highly depend on the efficiency of the external pose estimation or semantic segmentation techniques. In addition to that, they are suffering misalignment issues. Thus, to address these issues, state-of-the-art approaches~\citep{sun2018beyond} take into account global features and then divide into parts or stripes. The advantage of using global features for local features representations is two folds: i)  does not suffer from misalignment caused by inaccurate bounding box detection, human pose changes and various human spatial distributions. ii) different channels of the global feature have different recognition patterns which increase the discriminative ability of the extracted feature  by paying weighted attention to different parts of human body.  Thus, we are focusing on the state-of-the-art methods that concentrates more on partitioning global features to form a local part based feature representations.

\subsection{Multiple Losses:}

Recently, there have been   efforts~\cite{zhou2019omni,luo2019bag,dai2019batch,shen2018end} to adopt multi-loss training strategies. More specifically, a combination of cross-entropy and triplet losses have proven to be effective to optimize ReID networks. The aim of this combination is to increase the discriminant power of the feature embedding by optimizing different objective functions. In~\cite{zhou2019omni}, an omni-scale feature learning scheme is designed to capture the  salient features at different scales that suggested optimizing ReID network with the combination of  cross-entropy and triplet losses for better performance. To achieve similar objective of having multi-scale feature learning, Niki et al.~\cite{martinel2020deep} proposed a pyramid-inspired deep ReID architecture where multi-loss functions combined with curriculum learning strategy to optimize the network.  In~\cite{zheng2019re}, an attention-driven Siamese learning architecture is designed to integrate attention and attention consistency by jointly optimizing the cross-entropy loss, the identification attention loss, and the Siamese attention loss. Chen at el.\cite{chen2019self} uses a reinforcement learning technique to quantify  the attention quality and provides a powerful supervisory signal to guide the learning process. Following the same trend, they use the combination of cross-entropy and triplet losses to optimize their proposed architecture. 

All of these ReID approaches above focus on improving the recognition accuracy without addressing the scalability issues to reduce computational costs. A few ReID approaches~\cite{wu2019distilled, hafner2018cross,liu2019adversarial,mekhazni2020unsupervised, fang2018perceptual, gong2020faster} seek to address the issues of computational complexity. Of these ReID approaches ~\cite{wu2019distilled,hafner2018cross}, few rely on distillation based approaches where knowledge is distilled from a deeper CNN (teacher model) to a lighter CNN (student model). Other approaches  ~\cite{liu2019adversarial,fang2018perceptual,gong2020faster} rely on hashing to learn binary representation instead of real-value feature for faster computation. In contrast to these approaches, our propose ReID approach relies on the pruning techniques that compresses the deep CNN with a marginal reduction of recognition accuracy. Additionally, we propose different design scenarios or pipelines for leveraging a pruning method during the deployment of a CNN for a target ReID application domain.


\section{Techniques for Pruning CNNs}
\label{S:2}

The objective of pruning is to remove unnecessary parameters from a neural network, while trying to maintain a comparable accuracy. Currently, pruning techniques operate on two different levels. First, techniques for weight-level pruning focus on pruning individual weights of a network. In contrast, techniques for channel-level pruning focus on removing all the parameters of the output and input channels of convolution layers. While weight pruning techniques can achieve high compression rate and good acceleration, its performance depends on a good sparse convolution algorithm which is unavailable, and does not perform well on all platforms. In this paper, we focus on channel pruning techniques which do not rely on other algorithms, and have been extensively studied in literature. This section presents a survey of channel pruning techniques, and summary of experimental result reported in the literature.

\subsection{Channel Pruning Taxonomy:}

Table 2 presents the main properties of different pruning techniques according to strategy used to reduce channels. In order to facilitate the analysis of different pruning methods, we also categorize techniques according to type of pruning criterion. In this table, "prune in one step" refers to techniques that prune the network one time and then fine-tune the network ~\citep{HaoLi, Entropy_Pruning, Redundant_Pruning}. "Prune iteratively" is a type of pruning that is done iteratively on a trained model that alternate between pruning and fine-tuning ~\citep{Molchanov}. Pruning by regularization is usually done by adding a regularization term to the original loss function in order to leave the pruning process for the optimization ~\citep{AutoBalanced, PlayAndPrune}. Pruning by minimizing the reconstruction error is a family of algorithms that tries to minimize the difference of outputs between the pruned and the original model. "Progressive pruning," while very similar to iterative pruning, differs in that it can start directly from a model that was not trained, and progressively prune it during training.

\begin{table*}[]
\centering
\label{prun_pro}
\caption{Main properties of different channel pruning techniques.}
\begin{adjustbox}{width=1\textwidth}

\begin{tabular}{|l||l|l|}
\hline
Strategy                                                      & Methods           & Criteria                                                                                                                                                                                                                    \\ \hline \hline
Prune in one step                           & L1 ~\citep{HaoLi}                & weights:  $ S_{j}=\sum \left |  w_{k} \right | $       
\\ \cline{2-3} 
                                                              & Redudant Channels ~\citep{Redundant_Pruning} & weights:  $SIM_C(\textbf{W}_i,\textbf{W}_j) = \frac{\textbf{W}_{i}\boldsymbol{\cdot} \textbf{W}_{j}}{\left \| \textbf{W}_{i} \left \| \boldsymbol{\cdot} \right \| \textbf{W}_{j} \right \|}$                               \\ \cline{2-3} 
                                                              & Entropy ~\citep{Entropy_Pruning}           & feature maps:  $E_{j} = -\sum_{a=1}^{m}(p_{a}log(p_{a}))$                                                                                                                                                                   \\ \hline
Prune iteratively                            & Taylor ~\citep{Molchanov}            & feature maps: $\left | \Delta C(\textbf{H}_{i,j}) \right | = \left | \frac{\delta C}{\delta \textbf{H}_{i,j}} \textbf{H}_{i,j}\right |$                                                                                     \\ \cline{2-3} 
                                                              & FPGM ~\citep{FPGM}              & weights: $\textbf{W}_{i,j^*} \in {argmin}_{j^* \in R^{n_{in} \times k * \times k}} \sum_{j' \in [1, n_{out}]} ||x - \textbf{W}_{i,j'}||_2$                                                                   \\ \hline
Prune iteratively with regularization        & Play and Prune ~\citep{PlayAndPrune}   & weights: $S{_{j}}=\sum \left |  w_{k} \right |$   \\ \cline{2-3}    
                                                              & Auto-Balance ~\citep{AutoBalanced}     & weights: $S{_{j}}=\sum \left |  w_{k} \right |$                                                                                                                                                                             \\ \hline
Prune iteratively,  min reconstruction error & ThiNet            & feature maps:  $\textbf{H}_{i+1,j} = \sum_{j=1}^{C} \sum_{k=1}^{K} \sum_{k=1}^{K} \textbf{W}_{i,j,k,k}*\textbf{H}_{i,j}$                                                                                                    \\ \cline{2-3} 
                                                              & Channel Pruning ~\citep{ChannelPruning}  & feature maps:   ${\underset{\beta,\textbf{W}}{\arg \min} \tfrac{1}{2N}\left \| \textbf{H}_{i+1,j} - \sum_{j=1}^{n} \beta_{i,j} \textbf{H}_{i,j} \textbf{W}_{i,j} \right \|_{F}^{2}+ \lambda \left \| \beta \right \|_{1} }$ \\ \hline
                                            Prune progressively                  & PSFP ~\citep{Pro_Soft_Pruning}             & weights: $S{_{j}}=\sum \left |  w_{k} \right |_2$                                                                                                                                                                           \\ \hline
\end{tabular}
\end{adjustbox}
\end{table*}
 
One key challenge of pruning neural networks is selecting the pruning criteria. It should allow to discern the parameters that contribute to the accuracy and the ones that do not. Another challenge is finding an optimal pruning compression. This compression ratio is essential to find a compromise between the reduction of complexity for model and the loss of accuracy. Finally, one challenge is the retraining and pruning schedule of the model. Punning can be performed in one iteration but the damage caused to the network may be considerable. On the counterpart, we could prune and retrain iteratively to reduces this damage at each iteration, yet but this will take longer to apply.  The retraining of the pruned network may also cause the model to overfit or get caught in local minimums.

\subsection{Description of methods:}

This subsection presents different pruning algorithms for each pruning family in the taxonomy. To ease our notation, we refer to a convolution tensor as $\textbf{W}$ with $\textbf{W} \in R^{n_\text{out} \times n_\text{in} \times k \times k}$, $n_{in}$ the number of input channels, $n_{out}$ the number of output channels, $k$ the kernel size. An output channel tensor $i$ is then defined as $\textbf{W}_{i}$, and an individual weight is defined as $\textbf{w}$. For feature map, $\textbf{H}$ represents an output of a convolution layer,$\textbf{H}_i$ then represent the output channel of a feature map. For ease of notation, we do not mention the layer index unless necessary, therefore $\textbf{W}$ or $\textbf{H}$ can be any convolution layer or feature map at any index.

\subsubsection{Criteria based on weights:}

The L1 ~\citep{HaoLi} pruning algorithm is layer-by-layer method which means it will prune the network one layer at a time. This algorithm's pruning criteria is simple and could be implemented using Algorithm 1.

\begin{algorithm}[ht]
\caption{L1 Pruning}\label{alg:L1_Pruning}
\begin{algorithmic}[1]
\State {\textbf{Input:} training data: \textbf{X}} 
\State {\textbf{Input:} pruning rate: $P_{i}$}
\State {\textbf{Input:} the model with parameters \textbf{M} = \{$\textbf{M}^{(i)}$,$0$ $\leq$ $i$ $\leq$ $L$\}}
\State {\hspace{8pt}Initialize the model parameter \textbf{M}} 
\State {\hspace{8pt}\textbf{for} each convolution layer  \textbf{do}} 
\State {\hspace{16pt}Calculate $l_{1}-$norm for each channel}
\State {\hspace{16pt}Select the $N$ lowest $l_{1}-$norm depending on the pruning rate}
\State {\hspace{16pt}Remove the $N$ selected channels}
\State {\hspace{8pt}\textbf{end for}}
\State {\hspace{8pt}Re-Train the pruned network}
\State {\textbf{Output:} Compact model with parameters \textbf{$M'$}}
\end{algorithmic}
\end{algorithm}

The retraining could be done in two different ways:
\begin{enumerate}
\item Prune once over multiple layers and retrain (more adapted for resilient layers)
\item Prune channels one by one and retrain each time (more adapted for layers that are less resilient)
\end{enumerate}
For this algorithm, we chose to mix the two retraining methods to come up with pruning N channels before retraining.

The second weight method that will be presented is the redundant channel pruning ~\citep{Redundant_Pruning}. This method's idea is to pruned channels that are similar to the ones that are kept. To do so, the authors proposed to regroup each channel of a layer in $n_{f}$ clusters depending on a similarity score being higher than a preassigned threshold $\tau$. To determine the similarity between these channels, the authors proposed to use the cosine similarity between the weights of the channels.

\begin{equation}
\label{eq:clus_sim}
\begin{aligned}
\overline{SIM_C}(C_a, C_b) = \frac{\sum_{\textbf{W}_i\in C_a, \textbf{W}_j\in C_b}SIM_C(\textbf{W}_i, \textbf{W}_j)}{|C_a| \times |C_b|} > \tau \\
a,b = 1,...,n_{out}; a \neq b; i =1,... |C_a|; j = 1,... |C_b|; i \neq j
\end{aligned}
\end{equation}

With the calculation of $SIM_C$ of two output channel given below: 
\begin{equation}
\label{eq:channel_sim}
SIM_C(\textbf{W}_i,\textbf{W}_j) = \frac{\textbf{W}_{i}\boldsymbol{\cdot} \textbf{W}_{j}}{\left \| \textbf{W}_{i} \left \| \boldsymbol{\cdot} \right \| \textbf{W}_{j} \right \|}
\end{equation}
\noindent
Equ. \ref{eq:channel_sim} gives us the ability to determine the similarity between two channels by calculating the cosine of the angle between two vectors of dimension n.
The pruning of one specific layer could be done in 2 steps:
\begin{enumerate}
    \item Group the channels in the same cluster if $cos(\theta)$ from Equ. \ref{eq:clus_sim} is above the threshold $\tau$
    \item Randomly sample one channel in each cluster and pruned the remaining ones of each cluster.
\end{enumerate}

\noindent
The threshold $\tau$ \space acts as the compression ratio in this pruning algorithm where a low threshold means a high compression rate and vice versa (see Algorithm 2). 

\begin{algorithm}[ht]
\caption{$SIM_C$ \space Pruning}\label{alg:SIMC_Pruning}
\begin{algorithmic}[1]
\State {\textbf{Input:} training data: \textbf{X}} 
\State {\textbf{Input:} pruning threshold $\tau$}
\State {\textbf{Input:} the model with parameters \textbf{M} = \{$\textbf{M}^{(i)}$,$0$ $\leq$ $i$ $\leq$ $L$\}}
\State {\hspace{8pt}Initialize the model parameter \textbf{M}} 
\State {\hspace{8pt}\textbf{for} each convolution layer  \textbf{do}} 
\State {\hspace{16pt}Calculate the similarity score for each pair of channel}
\State {\hspace{16pt}Separate the channels into two clusters based on threshold $\tau$}
\State {\hspace{16pt}Select the $N$ lowest $l_{1}-$norm depending on the pruning rate}
\State {\hspace{16pt}Randomly sample one channel in each cluster and pruned the remaining ones of each cluster.}
\State {\hspace{8pt}\textbf{end for}}
\State {\hspace{8pt}Re-Train the pruned network}
\State {\textbf{Output:} Compact model with parameters \textbf{$M'$}}
\end{algorithmic}
\end{algorithm}

The third weight-based method that will be presented is the Auto-Balanced pruning ~\citep{AutoBalanced} that uses the same pruning criteria as the L1 algorithm which is a L1 norm of weight kernels to determine the ranking of the channels. But this method adds a regularization term during the training to transfer the representational capacity of the channels we want to prune to the remaining ones. In order to calculate this transfer of representational capacity, the authors proposed to separate the channels in two subsets at the beginning of each pruning iteration.  In order to assign the channels to their subset, the authors used the L1 norm of the weights of the channels. The $vec$ function is used to flatten the weight matrix into a vector and $\textbf{M}_{i,j}$ the metric measuring the importance of. Here, we use the notation of $\textbf{W}_{i,j}$ with $i$ represent the layer index and j the output channel index.

\begin{equation}
\label{eq:L1_Weights}
\textbf{M}_{i,j} = \left \| vec(\textbf{W}_{i,j})) \right \|_{1}
\end{equation}

\noindent
Once the L1 score has been calculated for each channel, they are then assigned to one of the subsets depending on the threshold $\theta$ which is fixed depending on the desired number of remaining channels per layer.
\begin{equation}
\label{eq:AB_Assign_P}
\textbf{M}_{i,j} > \theta_{i}\  \forall \textbf{W}_{i,j} \in \textbf{P}_{i}
\end{equation}
\noindent
\begin{equation}
\label{eq:AB_Assign_R}
\textbf{M}_{i,j} < \theta_{i}\  \forall \textbf{W}_{i,j} \in \textbf{R}_{i}
\end{equation}
\noindent
The channels in subset R (remaining) and subset P (to pruned) are then adjusted with an L2 regularisation term. The following equations are used to calculate this L2 adjustment factor:
\begin{equation}
\label{eq:AB_Lambda}
\lambda _{i,j} = \left\{\begin{matrix}
1+log \frac{\theta _{i}}{\textbf{M}_{i,j}+\varepsilon } \ \text{if \space} \textbf{W}_{i,j} \in \textbf{P}_{i}\\\\

-1-log \frac{\textbf{M}_{i,j}}{\theta _{i}+\varepsilon } \ \text{if \space} \textbf{W}_{i,j} \in \textbf{R}_{i}
\end{matrix}\right.
\end{equation}
\vspace{-.9cm}
\noindent

\begin{equation}
\label{eq:AB_SPi}
S(\textbf{P}_{i})= \sum_{\textbf{W}_{i,j} \in \textbf{P}_{i}} \lambda _{i,j} \left \| vec(\textbf{W}_{i,j}) \right \|_{2}^{2},  \ \ \ \
S(\textbf{R}_{i})= \sum_{\textbf{W}_{i,j} \in \textbf{R}_{i}} \lambda _{i,j} \left \| vec(\textbf{W}_{i,j}) \right \|_{2}^{2}
\vspace{-2cm}
\end{equation}

\begin{equation}
\label{eq:AB_SP}
S(\textbf{P})= \sum_{i=1}^{n} S(\textbf{P}_{i}), \ \ \ \
S(\textbf{R})= \sum_{i=1}^{n} S(\textbf{R}_{i})
\end{equation}
\noindent

The cost function for training is changed with Equ. \ref{eq:AB_Loss} where $L_{0}$ represents the original cost function.
\begin{equation}
\label{eq:AB_Loss}
L = L_{0} + \alpha S(\textbf{P})+ \tau S(\textbf{R})
\end{equation}
\noindent
\begin{equation}
\label{eq:AB_tau}
\tau = -\alpha \tfrac{S(\textbf{P})}{S(\textbf{R})}
\end{equation}
\noindent
This enables the model to penalize the weak channels and stimulate the strong ones. This method adds two hyper-parameters in the training which are $\alpha$ \space and $r$. $\alpha$ \space is the regularization factor and the vector $r$ is the target of remaining channels in each layer (see Algorithm 3). 

\begin{algorithm}[ht]
\caption{Auto-Balance}\label{alg:Auto_Balance}
\begin{algorithmic}[1]
\State {\textbf{Input:} training data: \textbf{X}} 
\State {\textbf{Input:} pruning threshold $\tau$, $\alpha$, $r$}
\State {\textbf{Input:} the model with parameters \textbf{M} = \{$\textbf{M}^{(i)}$,$0$ $\leq$ $i$ $\leq$ $L$\}}
\State {\hspace{8pt}Initialize the model parameter \textbf{M}} 
\State {\hspace{8pt}\textbf{for} each convolution layer  \textbf{do}} 
\State {\hspace{16pt}Divide the channels into $R$(remain) subset and $P$(Prune) subset using $l_{1}-$norm}
\State {\hspace{16pt}Optimize the Equ.\ref{eq:AB_tau}}
\State {\hspace{8pt}\textbf{end for}}
\State {\textbf{Output:} Compact model with parameters \textbf{$M'$}}
\end{algorithmic}
\end{algorithm}

The last weight-based method is the Progressive Soft Pruning ~\citep{Pro_Soft_Pruning} where their pruning criterion is the same as the L1 method (L2 norm of the weights). The main difference with this method is they proposed an interesting pruning scheme that allows pruning during the fine-tuning step. The authors proposed to use soft pruning which means instead of removing the channels during the pruning, they set the weights to 0, and allow these channels to be updated during the retraining phase. This pruning scheme is very interesting since the model keeps its original dimension during the retraining phase. The authors also proposed to add a progressive pruning scheme where at each pruning iteration, the compression ratio is increased in order to get a shallower network. Once these iterations of pruning and retraining are completed they do a last channel ranking using a pruning criteria and they discard the lowest channels depending on the compression ratio. Their pseudo-code for the progressive soft pruning scheme can be viewed in algorithm \ref{alg:PSFP}. In the article, they used the L1 or L2 norm of the weights as a pruning criteria which means this method could be categorized as a weight-based method.

\begin{algorithm}[t]
\caption{Algorithm Description of PSFP }\label{alg:PSFP}
\begin{algorithmic}[1]
\State {\textbf{Input:} training data: \textbf{X}} 
\State {\textbf{Input:} pruning rate: $P_{i}$, pruning rate decay $D$}
\State {\textbf{Input:} the model with parameters \textbf{M} = \{$\textbf{M}^{(i)}$,$0$ $\leq$ $i$ $\leq$ $L$\}}
\State {\hspace{8pt}Initialize the model parameter \textbf{M}} 
\State {\hspace{8pt}\textbf{for} $epoch = 1;$ $epoch$ $\leq$ $N_{epoch};$ $epoch++$ \textbf{do}} 
\State {\hspace{16pt}Update the model paramters \textbf{M} based on $X$} 
\State {\hspace{16pt}\textbf{for} $i = 1;$ $i$ $\leq$ $L;$ $i++$ \textbf{do}} 
\State {\hspace{32pt}Calculate the $l_{2}-$norm for each channel} 
\State {\hspace{32pt}Calculate the pruning rate $P'$ at this epoch using $P_{i}$ and $D$}
\State {\hspace{32pt}Select the $N$ lowest $l_{2}-$norm depending on the pruning rate}
\State {\hspace{32pt}Zeroize the weights $W$ of the selected channels}
\State {\hspace{16pt}\textbf{end for}}
\State {\hspace{8pt}\textbf{end for}}
\State {\hspace{8pt}Obtain the compact model with parameters \textbf{M'} from \textbf{M}}
\State {\textbf{Output:} Compact model with parameters \textbf{$M'$}}
\end{algorithmic}
\end{algorithm}
\noindent
The $L$ represents the number of layers in the model, $i$ represents the layer number, $W$ represents the weights of a channel and $N$ is the number of channels to prune. The pruning rate $P'$ is calculated at each epoch using the pruning rate goal $P_{i}$  for the corresponding layer $i$ and the pruning rate decay $D$. To calculate the pruning rate, we can use Equ. \ref{eq:Pruning_Rate_PSFP}
\begin{equation}
P_{i}^{'} = a \cdot e^{-k \cdot epoch}+b
\label{eq:Pruning_Rate_PSFP}
\end{equation}
\noindent
The $a$,$b$ and $k$ values can be calculated by solving the Equ. \ref{eq:Pruning_PSFP_Equ.}
\begin{equation}
\left\{\begin{array}{l}
0 = a+b
\\
\frac{P_{i}}{4} = e^{-k \cdot N_{epoch} \cdot D} + b
\\ 
\end{array}\right.
\label{eq:Pruning_PSFP_Equ.}
\end{equation}
FPGM ~\citep{FPGM} is a new technique that focuses on using geometric median to prune away output channels. A geometric median is defined as: given a set of $n$ points $A = [a_1, a_2, ...,a_n]$ with $a_i \in R^d$, find a point $x^* \in R^d$ that minimizes the sum of the Euclidean distances to them:

\begin{equation}
\label{eq:geometric_median}
x^* \in \operatorname*{argmin}_{x \in R^d} f(x) \text{  where }  f(x)  \triangleq \sum_{i \in [1,n]} || x - a_i||_2
\end{equation}

Using the Equ. \ref{eq:geometric_median}, a geometric median $F_i^{GM}$ for all the filters of a layer $i$ can be found:

\begin{equation}
\label{eq:gemeotric_distance_layer}
\begin{aligned}
\textbf{W}_i^{GM} \in \operatorname*{argmin}_{x \in R^{n_{out} \times k * \times k}} g(x) \\
\text{where \space} g(x) \triangleq \sum_{j' \in [1, n_{in}]} ||x - \textbf{W}_{i,j'}||_2
\end{aligned}
\end{equation}

In order to select, non-important output channels, the author proposed to find the channels that have the same, or similar value of $\textbf{W}_i^{GM}$ which translates to:

\begin{equation}
\label{eq:gemeotric_distance_filter}
\begin{aligned}
\textbf{W}_{i,j^*} \in \operatorname*{argmin}_{x \in R^{n_{out} \times k * \times k}} ||\textbf{W}_{i,j^*} - \textbf{W}_{i}^{GM}||_2
\end{aligned}
\end{equation}

Since geometric median is a non-trivial problem, it's quite computationally intensive, therefore the authors propose to relax the problem by assuming that:

\begin{equation}
\label{eq:gemeotric_relax}
\begin{aligned}
 ||\textbf{W}_{i,j^*} - \textbf{W}_{i}^{GM}||_2 = 0
\end{aligned}
\end{equation}

This transforms the Equ. \ref{eq:gemeotric_distance_layer} to:
\begin{equation}
\label{eq:gemeotric_distance_final}
\begin{aligned}
\textbf{W}_{i,j^*} \in \operatorname*{argmin}_{j^* \in R^{n_{in} \times k * \times k}} \sum_{j' \in [1, n_{out}]} ||x - \textbf{W}_{i,j'}||_2 \\
&=  \operatorname*{argmin}_{j^* \in R^{n_{in} \times k * \times k}} g(x)
\end{aligned}
\end{equation}
The algorithm of FPGM is summarized in the Algorithm \ref{alg:FPGM}.

\begin{algorithm}[t]
\label{alg:FPGM}
\caption{Algorithm Description of FPGM }\label{alg:FPGM}
\begin{algorithmic}[1]
\State {\textbf{Input:} training data: \textbf{X}} 
\State {\textbf{Input:} pruning rate: $P$}
\State {\textbf{Input:} the model with parameters \textbf{M} = \{$\textbf{M}^{(i)}$,$0$ $\leq$ $i$ $\leq$ $L$\}}
\State {\hspace{8pt}Initialize the model parameter \textbf{M}} 
\State {\hspace{8pt}\textbf{for} $epoch = 1;$ $epoch$ $\leq$ $N_{epoch};$ $epoch++$ \textbf{do}} 
\State {\hspace{16pt}Update the model parameters \textbf{M} based on $X$} 
\State {\hspace{16pt}\textbf{for} $i = 1;$ $i$ $\leq$ $L;$ $i++$ \textbf{do}} 
\State {\hspace{32pt}Select the $n_{out} \times P$ of $W_i$ channels that satisfy Equ. \ref{eq:gemeotric_distance_final}}
\State {\hspace{32pt}Set the selected channels to zero}
\State {\hspace{16pt}\textbf{end for}}
\State {\hspace{8pt}\textbf{end for}}
\State {\hspace{8pt}Obtain the compact model with parameters \textbf{M'} from \textbf{M}}
\State {\textbf{Output:} Compact model with parameters \textbf{$M'$}}
\end{algorithmic}
\end{algorithm}

Play And Prune ~\citep{PlayAndPrune} is an adaptive output channel pruning technique, that, instead of focusing on a criterion, tries to find an optimal number of output channels that can be pruned away given an error tolerance rate. This technique is min-max game of two modules, The Adaptive Filter Pruning (AFP) module and the Pruning Rate Controller (PRC). The goal of the AFP is to minimize the number of output channels in the model while the PRC tries to maximize the accuracy of the remaining set of output channels.
This technique considers a model $M$ can be partitioned into two sets of important channels $\textbf{I}$ and unimportant channels $\textbf{U}$.
\begin{equation}
    U_{i} = \sigma_{\text{top \space} \alpha\%}(sort(\{|\textbf{W}_1|, |\textbf{W}_2|, ...| \textbf{W}_{n_{out}}|\}))
\end{equation}
$U_{i}$ represents all the unimportant channels of a layer $i$. It is selected by a selecting $\alpha\%$ channels of the result of the sort operation on the L1 norm of each output channels. Once an $U_{i}$ is selected, the authors propose to add an additional penalty to the original loss function in order to prune without loss of accuracy while helping the pruning process. The original loss function would then become:
\begin{equation}
\label{eq:opti_remove_filters}
\begin{aligned}
\Theta = \operatorname*{argmin}_{\Theta} (C(\Theta) + \lambda_A||\textbf{U}||_1)
\end{aligned}
\end{equation}
where $C(\Theta)$ is the original cost function to optimize the original model parameters, and $\lambda_A$ is the $L_1$ regularization term. While this optimization help pushing the channels to have zero sum of absolute weights, it can take some epochs, therefore the authors propose an adaptive weight threshold ($\textbf{W}_i$) for each layer $i$. Any channels with L1 norm below this threshold will be removed. While this value is given by the PRC, for the first epoch, it found by using a binary search on the histogram of sum of absolute weights. The AFP minimizes the number of output channels in the model using Equ. \ref{eq:opti_remove_filters}. The AFP can be summarized in Algorithm \ref{alg:AFP}, and the loss function of AFP can be written as:
\begin{equation}
\label{eq:AFP}
\begin{aligned}
\Theta' = \operatorname*{\sigma}_{\#w \in \Theta'}[P(\operatorname*{argmin}_{\Theta'} (C(\Theta) + \lambda_A||\textbf{U}||_1))]
\end{aligned}
\end{equation}
For the PRC, the adaptive threshold $W_A$ is updated as follows:
\begin{equation}
\label{eq:W_A}
    \textbf{W}_A = \delta_w  \times T_r \times \textbf{W}'_A
\end{equation}

With $\delta_w$ the constant used to increase or decrease the pruning rate. $T_r$ is calculated as follow:
\begin{equation}
T_r = \left\{\begin{array}{l}
C(\#w) - (\xi - \varepsilon) :C(\#w) - (\xi - \varepsilon) > 0
\\ 
0\text{\space \space: Otherwise}
\end{array}\right.
\end{equation}
where $\xi$ is the accuracy of the unpruned network, $\varepsilon$ is the tolerance error, $C(\#w)$ is the accuracy of the model with the remaining filter $\#w$. The regularization constant is also computed as follows:
\begin{equation}
\lambda_A = \left\{\begin{array}{l}
C(\#w) - (\xi - \varepsilon) \times \lambda :C(\#w) - (\xi - \varepsilon) > 0
\\ 
0\text{\space \space: Otherwise}
\end{array}\right.
\end{equation}
with $\lambda$ the initial regularization constant. By alternating between the AFP and the PRC, the authors propose a system that prune at each epoch in an adaptive and iterative way (see Algorithm 6). 

\begin{algorithm}[ht]
\caption{AFP}\label{alg:AFP}
\begin{algorithmic}[1]
\State {\textbf{Input:} training data: \textbf{X}} 
\State {\textbf{Input:} the model with parameters \textbf{M} = \{$\textbf{M}^{(i)}$,$0$ $\leq$ $i$ $\leq$ $L$\}}
\State {\hspace{8pt}Initialize the model parameter \textbf{M}} 
\State {\hspace{8pt}\textbf{for} each convolution layer  \textbf{do}} 
\State {\hspace{16pt}Select an $\alpha\%$ of output channels with the lowest $l_{1}-$norm}
\State {\hspace{16pt}Separate the output channels into $U$(Unimportant) and $I$(Important) subsets}
\State {\hspace{16pt}Perform Equ. \ref{eq:opti_remove_filters} with $\lambda_A$ Agiven by PRC}
\State {\hspace{16pt}Remove unimportant channels using the threshold $W_A$ given by PRC}
\State {\hspace{8pt}\textbf{end for}}
\State {\textbf{Output:} Compact model with parameters \textbf{$M'$}}
\end{algorithmic}
\end{algorithm}

\subsubsection{Criteria based on feature maps:}

In the channel-based approach, we are going to present 3 algorithms which are ThiNet ~\citep{ThiNet}, Channel Pruning ~\citep{ChannelPruning} and Entropy Pruning ~\citep{Entropy_Pruning}. The first two algorithms have the same idea behind their pruning algorithm which is minimizing the difference in the activation maps but they diverge with their  minimization technique. ThiNet's goal is to find a subset of channels that minimise the difference in the output at layer i+1 (feature map). 

ThiNet uses greedy algorithm to find which subset of channels to eliminate and keep the input at layer i+2 almost intact. To find the subset of channels to prune, the authors proposed to use a greedy algorithm where they compute the value for each channel in a layer and assign the lowest value to the subset. They repeat this method until our pruning subset respects the defined compression ratio. To calculate the input of the feature map in layer i+2, we can use the Equ. \ref{eq:Feature_Map_ThiNet}. 
\begin{equation}
\label{eq:Feature_Map_ThiNet}
\textbf{H}_{i+1,j} = \sum_{j=1}^{C} \sum_{k=1}^{K} \sum_{k=1}^{K} \textbf{W}_{i,j,k,k}*\textbf{H}_{i,j}
\end{equation}
\noindent
Where $i$ represents the layer, $j$ the channel index and $k$ the kernel size of the channel. To compute the value of a channel, the authors proposed to used Equ. \ref{eq:Value_ThiNet}
\begin{equation}
\label{eq:Value_ThiNet}
\sum_{i=1}^{m} \hat{x}_{i,j}^{2} , 
\end{equation}
\noindent
where $\hat{x}$ is equal to $\textbf{W}_{i+1,j}$ in Equ. \ref{eq:Feature_Map_ThiNet}. This greedy method is repeated for each layer needed to be pruned in the model.

The Channel Pruning method also as the goal to minimize the difference in the output (feature map) but their method is to find a subset of channels with a LASSO regression.
\begin{equation}
\label{eq:Channel_LASSO}
\underset{\substack{\\\left \| \beta  \right \|_{0} \leq n' \\\\ 0 \leq n' \leq n }}{\underset{\beta,\textbf{W}}{arg min} \tfrac{1}{2N}\left \| \textbf{H}_{i+1,j} - \sum_{j=1}^{n} \beta_{i,j} \textbf{H}_{i,j} \textbf{W}_{i,j} \right \|_{F}^{2}+ \lambda \left \| \beta \right \|_{1} }
\end{equation}
\noindent
$\beta$  represents channel mask that decides whether the channel is pruned or not. If $\beta$ is zero then the channel is no longer useful. The compression ratio is defined with $\lambda$.  The $n$ represents the number of channels and $n'$ represents the number of remaining channels. During the pruning iterations, the $W$ in Equ. \ref{eq:Channel_LASSO} is fixed which leaves us with only one variable to minimize which is $\beta$. The LASSO regression is used to find this $\beta$  mask that minimizes the difference in the output. As in the ThiNet method, this method also requires to redo these steps for every layer needed to be pruned.

The entropy pruning ~\citep{Entropy_Pruning} method is also a layer by layer algorithm but instead of trying to minimize the difference in the output like the two channels methods above, they used a different criteria based on the entropy of the feature maps produced by the channels. The idea behind their criteria is that a low entropy in the feature maps of a channel will most likely be less important in the decision of the network. With the entropy criterion defined as:

\begin{equation}
\label{eq:Entropy}
E_{j} = -\sum_{a=1}^{m}(p_{a}log(p_{a}))
\end{equation}

Pruning for layer $i$ is done according to Algorithm 7. 

\begin{algorithm}[ht]
\caption{Entropy Pruning}\label{alg:Entropy_pruning}
\begin{algorithmic}[1]
\State {\textbf{Input:} training data: \textbf{X}} 
\State {\textbf{Input:} Pruning rate: \textbf{P}} 
\State {\textbf{Input:} the model with parameters \textbf{M} = \{$\textbf{M}^{(i)}$,$0$ $\leq$ $i$ $\leq$ $L$\}}
\State {\hspace{8pt}Initialize the model parameter \textbf{M}} 
\State {\hspace{8pt}Run all the data through the network and collect features for each convolution layer}
\State {\hspace{8pt}\textbf{for} each convolution layer  \textbf{do}} 
\State {\hspace{16pt}Convert the activation maps into a vector of dimension $n_{out}$ (number of channels) using a global average pooling.}
\State {\hspace{16pt}For each channel $j$ divide the distribution into $m$ bins and calculate the entropy using \ref{eq:Entropy}}
\State {\hspace{16pt}Remove channels with the lowest entropy value according to the pruning rate}
\State {\hspace{8pt}\textbf{end for}}
\State {\textbf{Output:} Compact model with parameters \textbf{$M'$}}
\end{algorithmic}
\end{algorithm}

Taylor\textquoteright s ~\citep{Molchanov} pruning  algorithm seeks to minimize the cost function. It approximates the change in this function if the channel is pruned according to:
\begin{equation}
\label{eq:Molchanov}
\left | \Delta C(\textbf{H}_{i,j}) \right | = \left | C(D,\textbf{H}_{i,j}=0) - C(D,\textbf{H}_{i,j})\right |
\end{equation}
\noindent
Where $C$ represents the cost function and  $D$ the dataset. $C(D,\textbf{H}_{i,j}=0)$ is the cost value if channel $\textbf{H}_{i,j}$ is pruned. 
The idea is to find a subset of channels $H_{i,j}$ to prune while minimizing the difference with the original Cost Function were these channels were used. This is represented in the equation by calculating the difference between cost function with the channels excluded and the cost function with the channels included.
Using a Taylor expansion to solve this minimization the authors found that the difference in the cost function with the channels pruned could be approximated with the activation (feature map) and the gradient of the channel which can be calculated during back-propagation.
\begin{equation}
\label{eq:Mol_Estimation}
\left | \Delta C(\textbf{H}_{i,j}) \right | = \left | \frac{\delta C}{\delta \textbf{H}_{i,j}} \textbf{H}_{i,j}\right |
\end{equation}
\noindent
Each channel ranking value is normalized using a l2-normalization. This normalization is done on each layer individually in order to facilitate the comparison between layers since this method ranks channels across all layers (see Algorithm 8):

\begin{algorithm}[ht]
\caption{Taylor Pruning}\label{alg:Taylor_pruning}
\begin{algorithmic}[1]
\State {\textbf{Input:} training data: \textbf{X}} 
\State {\textbf{Input:} Stopping condition}
\State {\textbf{Input:} the model with parameters \textbf{M} = \{$\textbf{M}^{(i)}$,$0$ $\leq$ $i$ $\leq$ $L$\}}
\State {\hspace{8pt}Initialize the model parameter \textbf{M}} 
\State {\hspace{8pt}\textbf{for} $epoch = 1;$ $epoch$ $\leq$ $N_{epoch};$ $epoch++$ \textbf{do}} 
\State {\hspace{16pt}Evaluate the importance of output channels using $X$ and Equ. \ref{eq:Mol_Estimation}} 
\State {\hspace{16pt}\textbf{for} each convolution layer  \textbf{do}} 
\State {\hspace{32pt}Remove channel with the least importance} 
\State {\hspace{16pt}\textbf{end for}}
\State {\hspace{16pt}Fine-tune the pruned network}
\State {\hspace{16pt}\textbf{if} Stopping condition is \textbf{True}}
\State {\hspace{32pt}\textbf{Break}}
\State {\hspace{8pt}\textbf{end for}}
\State {\hspace{8pt}Obtain the compact model with parameters \textbf{M'} from \textbf{M}}
\State {\textbf{Output:} Compact model with parameters \textbf{$M'$}}
\end{algorithmic}
\end{algorithm}

\subsubsection{Output-Based}

The second output-based method is the Neuron Importance Score Propagation (NISP) ~\citep{NISP} where the pruning is done by back propagating channel scores across the model to determine which ones to prune. The intuition behind their idea is to use a feature ranking method on the last layer before the classification since this layer is the one to play a more significant role in our application. Once every feature as an associated score, the authors proposed to back-propagate that score into the network to have an importance score for each channel in the network. The importance score is then used to determine which channels we pruned and which one we retain by using a predefined compression ratio for each layer in the model. 

\noindent
The score is back propagated using Equ. \ref{eq:NISP_Score}.
\begin{equation}
\label{eq:NISP_Score}
S_{i,j}=\sum_{i}|W_{i,j}^{(k+1)}|S_{i+1,j}
\end{equation}
\noindent
Where $W$ is the weights, $j$ is the neuron or channel, $i$ is the layer and $k$ is the number of connections from that neuron to the next layer. This equation represents a weighted sum of the scores in the subsequent layers.

\subsection{Critical analysis of pruning methods:}

The main difference between methods using a weight-based criteria versus methods based on feature map criteria is that they are not dependent of a dataset since weight statistic do not depend on output of a CNN. Methods based on feature maps need a dataset in order to compute the output of convolution layers or its gradients.

The chosen criteria usually depend on the desire to simplify the pruning steps at the expense of a lower accuracy -- some more complex criteria that require more computations allow preserving a high level of accuracy. If training and pruning time is an issue, i.e., applications with design constraints, and that requires fast deployment, a simple criteria like L1 and L2 norm are more suitable. However, if there is no complexity constraints, some more complex pruning criteria, like the minimization in the difference of activation or cost functions, can outperform the simpler criteria, at the expense of more computations and time.

\begin{table*}[t!]
\centering
\caption{Performance of pruning techniques from the literature in terms of rank-1 accuracy and computational complexity (memory(M): number of parameters and time (T): time required for one forward pass). To ease comparison, we include the out results produced with ThiNet (channel pruning method).}
\begin{adjustbox}{width=0.8\textwidth}
\begin{tabular}{lcccccc}
\hline
\multicolumn{1}{|l|}{\textbf{Dataset}} & \multicolumn{6}{l|}{\textbf{ResNet56 trained on CIFAR10}} \\ \hline
\multicolumn{1}{|l|}{\textbf{Algorithm}} & \multicolumn{3}{c|}{\textbf{Original}} & \multicolumn{3}{c|}{\textbf{Pruned}} \\ \cline{2-7} 
\multicolumn{1}{|l|}{} & \multicolumn{1}{c|}{\textbf{rank-1}} & \multicolumn{1}{c|}{\textbf{T}} & \multicolumn{1}{c|}{\textbf{M}} & \multicolumn{1}{c|}{\textbf{rank-1}} & \multicolumn{1}{c|}{\textbf{T}} & \multicolumn{1}{c|}{\textbf{M}} \\ \hline
\multicolumn{1}{|l|}{L1  ~\citep{HaoLi}} & \multicolumn{1}{c|}{93.04} & \multicolumn{1}{c|}{0.125} & \multicolumn{1}{c|}{0.85} & \multicolumn{1}{c|}{93.06} & \multicolumn{1}{c|}{0.091} & \multicolumn{1}{c|}{0.73} \\ \hline
\multicolumn{1}{|l|}{Auto-Balanced  ~\citep{AutoBalanced}} & \multicolumn{1}{c|}{93.93} & \multicolumn{1}{c|}{0.142} & \multicolumn{1}{c|}{N/D} & \multicolumn{1}{c|}{92.94} & \multicolumn{1}{c|}{0.055} & \multicolumn{1}{c|}{N/D} \\ \hline
\multicolumn{1}{|l|}{Redundant channel   ~\citep{Redundant_Pruning}} & \multicolumn{1}{c|}{93.39} & \multicolumn{1}{c|}{0.125} & \multicolumn{1}{c|}{0.85} & \multicolumn{1}{c|}{93.12} & \multicolumn{1}{c|}{0.091} & \multicolumn{1}{c|}{0.65} \\ \hline
\multicolumn{1}{|l|}{Play and Prune   ~\citep{PlayAndPrune}} & \multicolumn{1}{c|}{93.39} & \multicolumn{1}{c|}{0.125} & \multicolumn{1}{c|}{0.85} & \multicolumn{1}{c|}{93.09} & \multicolumn{1}{c|}{0.039} & \multicolumn{1}{c|}{N/D} \\ \hline
\multicolumn{1}{|l|}{FPGM   ~\citep{FPGM}} & \multicolumn{1}{c|}{93.39} & \multicolumn{1}{c|}{0.125} & \multicolumn{1}{c|}{0.85} & \multicolumn{1}{c|}{92.73} & \multicolumn{1}{c|}{0.059} & \multicolumn{1}{c|}{N/D} \\ \hline
 & \multicolumn{1}{l}{} & \multicolumn{1}{l}{} & \multicolumn{1}{l}{} & \multicolumn{1}{l}{} & \multicolumn{1}{l}{} & \multicolumn{1}{l}{} \\ \hline
\multicolumn{1}{|l|}{\textbf{Dataset}} & \multicolumn{6}{l|}{\textbf{VGG16 trained on ImageNet}} \\ \hline
\multicolumn{1}{|l|}{\textbf{Algorithm}} & \multicolumn{3}{c|}{\textbf{Original}} & \multicolumn{3}{c|}{\textbf{Pruned}} \\ \cline{2-7} 
\multicolumn{1}{|l|}{} & \multicolumn{1}{c|}{\textbf{rank-1}} & \multicolumn{1}{c|}{\textbf{T}} & \multicolumn{1}{c|}{\textbf{M}} & \multicolumn{1}{c|}{\textbf{rank-1}} & \multicolumn{1}{c|}{\textbf{T}} & \multicolumn{1}{c|}{\textbf{M}} \\ \hline
\multicolumn{1}{|l|}{ThiNet  ~\citep{ThiNet}} & \multicolumn{1}{c|}{90.01} & \multicolumn{1}{c|}{30.94} & \multicolumn{1}{c|}{138.34} & \multicolumn{1}{c|}{89.41} & \multicolumn{1}{c|}{9.58} & \multicolumn{1}{c|}{131.44} \\ \hline
\multicolumn{1}{|l|}{Taylor  ~\citep{Molchanov}} & \multicolumn{1}{c|}{89.30} & \multicolumn{1}{c|}{30.96} & \multicolumn{1}{c|}{N/D} & \multicolumn{1}{c|}{87.06} & \multicolumn{1}{c|}{11.5} & \multicolumn{1}{c|}{N/D} \\ \hline
\multicolumn{1}{|l|}{HaoLi  ~\citep{HaoLi}} & \multicolumn{1}{c|}{90.01} & \multicolumn{1}{c|}{30.94} & \multicolumn{1}{c|}{138.34} & \multicolumn{1}{c|}{89.13} & \multicolumn{1}{c|}{9.58} & \multicolumn{1}{c|}{130.87} \\ \hline
\multicolumn{1}{|l|}{Channel Pruning  ~\citep{ChannelPruning}} & \multicolumn{1}{c|}{90.01} & \multicolumn{1}{c|}{30.94} & \multicolumn{1}{c|}{138.34} & \multicolumn{1}{c|}{88.10} & \multicolumn{1}{c|}{7.03} & \multicolumn{1}{c|}{131.44} \\ \hline
 & \multicolumn{1}{l}{} & \multicolumn{1}{l}{} & \multicolumn{1}{l}{} & \multicolumn{1}{l}{} & \multicolumn{1}{l}{} & \multicolumn{1}{l}{} \\ \hline
\multicolumn{1}{|l|}{\textbf{Dataset}} & \multicolumn{6}{l|}{\textbf{ResNet50 trained on ImageNet}} \\ \hline
\multicolumn{1}{|l|}{\textbf{Algorithm}} & \multicolumn{3}{c|}{\textbf{Original}} & \multicolumn{3}{c|}{\textbf{Pruned}} \\ \cline{2-7} 
\multicolumn{1}{|l|}{} & \multicolumn{1}{c|}{\textbf{rank-1}} & \multicolumn{1}{c|}{\textbf{T}} & \multicolumn{1}{c|}{\textbf{M}} & \multicolumn{1}{c|}{\textbf{rank-1}} & \multicolumn{1}{c|}{\textbf{T}} & \multicolumn{1}{c|}{\textbf{M}} \\ \hline
\multicolumn{1}{|l|}{Entropy  ~\citep{ Entropy_Pruning}} & \multicolumn{1}{c|}{72.88} & \multicolumn{1}{c|}{3.86} & \multicolumn{1}{c|}{25.56} & \multicolumn{1}{c|}{70.84} & \multicolumn{1}{c|}{2.52} & \multicolumn{1}{c|}{17.38} \\ \hline
\multicolumn{1}{|l|}{ThiNet  ~\citep{ThiNet}} & \multicolumn{1}{c|}{75.30} & \multicolumn{1}{c|}{7.72} & \multicolumn{1}{c|}{25.56} & \multicolumn{1}{c|}{72.03} & \multicolumn{1}{c|}{3.41} & \multicolumn{1}{c|}{138.00} \\ \hline
\multicolumn{1}{|l|}{FPGM  ~\citep{FPGM}} & \multicolumn{1}{c|}{75.30} & \multicolumn{1}{c|}{7.72} & \multicolumn{1}{c|}{25.56} & \multicolumn{1}{c|}{74.83} & \multicolumn{1}{c|}{3.58} & \multicolumn{1}{c|}{N/D} \\ \hline
\end{tabular}%
\end{adjustbox}
\label{Summary Article Results}
\end{table*}

Some of the techniques also differ in term how channels are pruned, some prune layer-by-layer~\citep{HaoLi, Pro_Soft_Pruning}, while other others prune across layers~\citep{Molchanov}. One of the differences between across layer and layer-by-layer pruning is the imbalance in term of pruning. An across layer pruning does not prune each layer evenly, and the method can possibly prune lower level layers more than higher level layers, and vice versa. Depending on the CNN architecture and pruning algorithm, pruning across layers may not yield the desired reduction. Layer-by-layer pruning can guarantee that all the layers will be pruned, and therefore undergo a more even reduction at each layer. 

Recently, some techniques~\citep{FPGM, Pro_Soft_Pruning} also adopt a new soft pruning approach. Soft pruning differs from hard pruning because it only resets pruned channels to zero instead of completely removing them. Therefore soft pruned channels have a chance to recover. These techniques have been shown to have achieved state-of-the-art performance.

Table~\ref{Summary Article Results} summarizes  a comparative experimental analysis of different pruning techniques. All the results reported in this table have been taken from the corresponding papers.
Experimental performance indicates that VGG16 processing can accelerate by up to 2.5 times at the expense of increased error of 1\% on the ImageNet dataset. Comparing L1 ~\citep{HaoLi} versus Auto-Balanced ~\citep{AutoBalanced} techniques, both based on a weight-based criteria, we observer that the  Auto-Balanced techniques can obtain higher compression ratios because of its regularization term. If we compare weight-based  and channel-based approaches, we observe that using channel-based provides a higher compression while maintaining similar accuracy. For the comparison between output and channel-based approaches, ThiNet ~\citep{ThiNet} outperforms  Taylor ~\citep{Molchanov} in accuracy and complexity.

\subsection{Design scenarios with pruning:}
\label{S:Prune_CNN_Design}

Most DL models for person ReID use pretraining, and then fine-tune the model to the task or target application domain. Pretraining is typically performed using a large-scale dataset in order to prime CNN parameters towards relevant optimization solutions. In many cases, CNNs are pretrained on ImageNet since this public dataset has a large amount of diverse training samples from different classes which improves the CNNs capacity to generalize. In person ReID, pretrained models have proven to be more successful than models that were trained from scratch directly on the task dataset. 

Once the model has been pretrained, the next step is fine-tuning to map the model's parameters from our pretraining source domain to our target application domain. It is crucial that the task dataset be similar to pretraining data. As described in~\citep{Finetune_Best}, the best fine-tuning practices depend on the size of the task training dataset, and the difference in data distribution between pretraining and task domain data. The authors propose to compute a similarity score between the pretraining and task datasets in order to guide the fine-tuning from one target domain to a another. They proposed measuring their similarity with the cosine distance and the maximum mean discrepancy (MMD). In particular, they proposed to average the feature embedding of each dataset and calculating the metrics between the two vectors. Given  these metrics and the number of samples per class in the target domain dataset, authors proposed to either train the whole network or freeze the feature extractor and fine-tune the classifier. We follow their guidelines to determine the layers to freeze and to fine-tune.

\begin{figure*}[t!]
    \centering    \includegraphics[width=12cm]{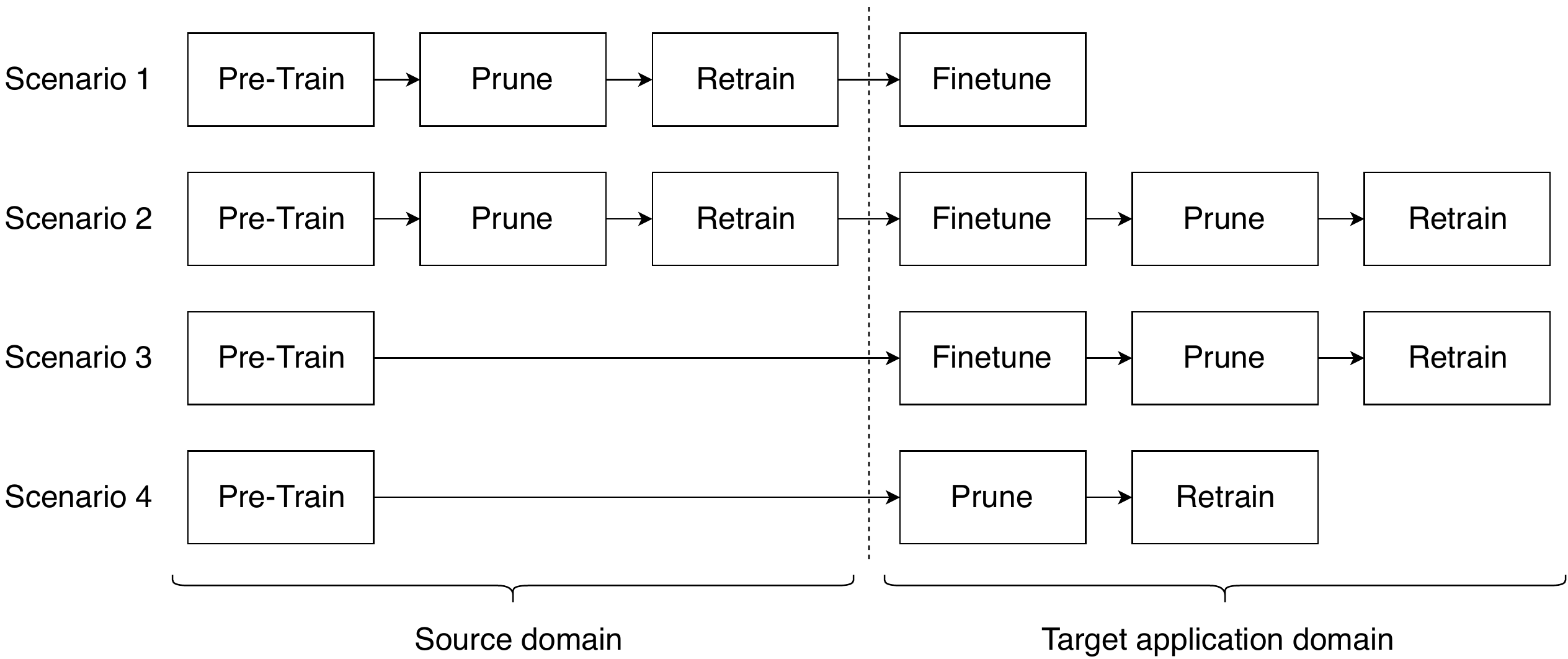}
    \caption{Scenarios for pruning and training a CNN.}
    \label{Pruning_Scenarios}
\end{figure*}

Pruning neural networks can be done in both main training phases -- pretraining and fine-tuning. We concluded that there are four possible scenarios for pruning (as shown in Figure \ref{Pruning_Scenarios}). The first scenario consists in pruning a CNN on the source pretraining dataset. The idea behind this scenario is to leverage a large-scale dataset to guide our selection of the more relevant and discriminant source domain channels. The second scenario consists in pruning on the source pretraining dataset, and then fine-tuning until our model provides a suitable performance, and then prune again on the target application dataset. This strategy allows removing additional channels that are not contributing to our task. The third scenario consists in only pruned on our task dataset after the fine-tuning on the target application dataset. The objective of this scenario is to accelerate the training time since pruning and retraining on a large-scale source domain dataset can be time consuming. Finally, the last scenario consists in pruning on the task dataset before doing the fine-tuning. This scenario goal also reduces design time of the model. In Section~\ref{S:4} we seek to determine the best scenario to reduce the computational  complexity of CNNs, while maintaining a comparable level of accuracy on our task.

The progressive soft pruning method is an interesting alternative since the model is pruned during  fine-tuning steps. This pruning scheme can reduce training effort since it combines the pruning, retraining, and fine-tuning into a single step. In Figure~\ref{Pruning_Scenarios}, progressive soft pruning would be represented by combining the prune and retrain process in one box for Scenario 1. For Scenario 2 and 3, the fine-tuning, pruning and retraining would be combined into one box. As for the Scenario 4, PSFP is not be applicable since the pruning, retrain, fine-tuning is one step, making it impossible to prune the network by ranking the channels with the target data, and then fine-tuning the network.


\section{Experimental Methodology}
\label{S:4}
In this section, we present the experimental methodology used to validate the pruning model. Our experiment is divided into two main parts. First, we experiment on a large-scale dataset, i.e. ImageNet, in order to find the best pruning methods using the same experimental protocol. The second part of these experiments will be to test the pruning algorithms on a person Re-ID problem to find the advantage of using a pruned model compared to smaller model. The following section will present the experimental methodology such as the datasets, the evaluation metrics and the experiments algorithm. The results for the pruning on the ImageNet dataset and ReID datasets will also be presented.

\subsection{Datasets:}  
Four publicly available datasets are considered for the experiments, namely Imagenet~\citep{ILSVRC12}, Market1501~\citep{market1501}, DukeMTMC-reID~\citep{DukeMTMC} and CUHK03-NP~\citep{cuhk03}. Imagenet, a large-scale dataset, is used as pre-trained dataset and rest of the other datasets (small-scale) are used for the experiments of person Re-IDs. 

\begin{itemize}

\item \textbf{ImageNet} (ILSVRC2012) \citep{ILSVRC12} is composed of two parts. The first part is used for training the model and the second part is used for validation/testing. There is 1.2M images for training and 50k for validation. The ILSVRC2012 dataset contains 1000 classes of natural images.

\item \textbf{Market-1501} \citep{market1501} is one of the largest public benchmark datasets for person Re-ID. It contains 1501 identities which are captured by six different cameras, and 32,668 pedestrian image bounding boxes obtained using the Deformable Part Models (DPM) pedestrian detector. Each person has 3.6 images on average at each viewpoint. The dataset is split into two parts: 750 identities are utilized for training and the remaining 751 identities are used for testing. We follow the official testing protocol where 3,368 query images are selected as probe set to find the correct match across 19,732 reference gallery images. 

\item \textbf{CUHK03-NP}~\citep{cuhk03} consists of 14,096 images of 1,467 identities. Each person is captured using two cameras on the CUHK campus, and has an average of 4.8 images in each camera. The dataset provides both manually labeled bounding boxes and DPM-detected bounding boxes. In this paper, both experimental results on ‘labeled’ and ‘detected’ data are presented. We follow the new training protocol proposed in \citep{zhong2017re}, similar to partitions of Market1501 dataset. The new protocol splits the dataset into training and testing sets, which consist of 767 and 700 identities, respectively. 

\item \textbf{DukeMTMC-reID}~\citep{DukeMTMC} is constructed from the multi-camera tracking dataset– DukeMTMC. It contains 1,812 identities. We follow the standard splitting protocol proposed in~\citep{zheng2017unlabeled} where 702 identities are used as the training set and the remaining 1,110 identities as the testing set. 

\end{itemize}

\subsection{Pruning methods:}
For our experiments, we compare five pruning methods in order to determined which technique gives the best compression ratio while maintaining a good performance on person ReID task. Our choice was based on the following criteria: article results, most of the families of the taxonomy are represented and the complexity for the ranking and the implementation. We selected \texttt{L1}~\citep{HaoLi} and \texttt{Entropy}~\citep{Entropy_Pruning} as they rely on the techniques that prune the network only one time and then fine-tune the network. Although \texttt{Taylor}~\citep{Molchanov} uses iterative pruning techniques, we chose this method for our experiments because of its theoretical explanation and requires single compression ratio. We choose to experiment with \texttt{Auto-Balanced} algorithm ~\citep{AutoBalanced} because pruning is done by adding regularization term to original loss function in order to leave the pruning process for the optimization. We've also decided to try the \texttt{Progressive Soft Pruning} ~\citep{Pro_Soft_Pruning} method since it directly prunes from scratch and progressively prune during training which is suitable test on our target operational domain.

\textbf{Implementation Details. }
For Triplet based ReID method, images are resized to $256 \times 128$ for all the datasets. For PCB~\citep{sun2018beyond} architectures, images are resized to $384 \times 128$ Like many state-of-art ReID approaches~\citep{hermans2017defense,chen2017beyond,geng2016deep,cheng2016person,liu2017end,sun2018beyond}, we use ResNet50 ~\citep{he2016deep} as the backbone architecture, where the final layer is removed to get a 2048 feature representation. We apply all the pruning method on the ResNet50 architecture. In order to be able to compare the four algorithms more easily, we decided to come up with a pruning schedule that would be similar for all the methods. First of all, we decided to prune around 5\% of the total number of channels at each iteration. 

For the layer-by-layer methods, we chose to use a single compression rate for every layer in order to simplify our experiments and our comparison between the methods. For each pruning iteration, we decided to use 1 epoch for the ranking of the channels and 4 epochs for retraining before moving to the next iteration.

This pruning schedule was used for every method on ImageNet in order to produced our pruned models that would be used in the person ReID experiments. We've discarded the pruning iterations where the accuracy was too low since there was no advantage of using these networks for our task. Once our pruning was done for every method, we retrained every model on ImageNet to regain the loss of accuracy cause by the pruning. Each of our pruned models was then fine-tune on the ReID datasets. We also fine-tuned pretrained ResNet18 and ResNet34 on these ReID datasets in order to compare the advantages of using pruned models compared to shallower networks.

\subsection{Performance metrics:} 

Following the common trend of evaluation~\citep{hermans2017defense,chen2017beyond,geng2016deep,cheng2016person,liu2017end}, we use the rank-01 accuracy of the cumulative matching characteristics (CMC), and the mean average precision (mAP) to evaluate the ReID accuracy.   he CMC represents the expectation of finding a correct match in the top n ranks. When multiple ground truth matches are available, then CMC cannot measure how well the gallery images are ranked. Thus, we also report the mAP scores.

As the state-of-the-art pruning methods~~\citep{HaoLi, Entropy_Pruning, ChannelPruning,Pro_Soft_Pruning}, the FLOPS's metric is used to calculate the model's complexity in terms of computational operations. To compare the different models during our experiments, we decided to calculate the number of FLOPS necessary to process one image through the model. We chose to compare the number of FLOPS since the processing time depends on the material used. The FLOPS is also a better metric than the number of pruned channels since a pruned channel at the beginning of the network will be reduced considerably more the total number of FLOPS than a later layer channel since the image dimension is reduced throughout the network. We also use the number of parameters metric to be able to compare models in terms of memory consumption to save the trained model. This metric was calculated by summing the number of weights needed throughout the model.


\section{Experimental Results and Discussion}
\label{S:4}

\subsection{Pruning on pre-training data:}

\begin{table}[!t]
\caption{Rank-1 accuracy and complexity (M: number of parameters and T: GFLOPS) of baseline and pruned ResNet CNNs on ImageNet dataset~\citep{imagenet_cvpr09}.}
\centering
\scalebox{1.0}{%
\begin{tabular}{|l|c|c|c|}
\hline
                & \multicolumn{3}{c|}{\textbf{Performance measures}} \\ \cline {2-4}
\textbf{Networks}  & \textbf{rank-1} & \textbf{T} & \textbf{M} \\ \hline \hline
ResNet50 & 76,01 & 6,32 & 23,48 \\ \hline
ResNet34 & 73,27 & 6,67 & 21,28 \\ \hline
ResNet18 & 69,64 & 3,09 & 11,12 \\ \hline \hline
L1 & 71,85 & 2,96 & 11,90 \\ \hline
Taylor & 71,65 & 3,21 & 12,09 \\ \hline
AutoBalanced & 71,97 & 2,96 & 11,90 \\ \hline
Entropy & 71,46 & 2,96 & 11,90 \\ \hline
PSFP & 71,57 & 2,96 & 11,90 \\ \hline
\end{tabular}%
}
\label{ImageNet Results}
\end{table}

Table~\ref{ImageNet Results} shows the performance of baseline and pruned ResNet CNNs (backbone ResNet50 and smaller ResNet 18 and ResNet34 networks) on the ImageNet dataset. Results in this table provide an indication of the benefits of pruning only on a large source domain pre-training dataset. There are few variations in results among the pruning techniques. The similarity in the results indicates that efficient pruning techniques rely on the availability of the large-scale datasets. Given the large-scale dataset in this experimental evaluation, ranking done by each technique is very similar even though the ranking metrics differ. Pruning results with ResNet50 very similar to the state of the art without pruning, although half the FLOPS are required. For example, using ThiNet \citep{ThiNet} provides a compressed network that yields rank-1 accuracy of around 72\% for half the FLOPS. Results with pruned networks provide  higher accuracy than the smaller ResNet18 while having similar computational complexity and memory consumption.


For person reID datasets, we attempt to preserve the same pruning compression ratio of 50\% for  comparison. This ratio is the highest compression level while minimizing the difference in the results between the baseline and the pruned models. We also prune same number of filters per layer, use the same number of pruning iteration, and same fine-tuning iteration, layer-by-layer. Across layers is not same, 50\% filter gone, 5\% filter at iteration, stopping condition is 50\% pruned away. 
 
Table~\ref{Market1501 Triplet Results} reports the results for Market-1501, DukeMTMC-reID and CUHK03-NP Re-IDs. The reported results are for all the \texttt{Scenario}. Taylor has higher FLOPS and a higher number of parameters than the other methods which would probably lead to a slower model and more consumption in terms of memory. Out of the 5 methods, the L1 method seems to be working the best by having the best or close to the best on the three datasets.

\begin{table*}[!t]
\centering
\caption{Accuracy (mAP and  rank-1:R-1) and complexity (M: Parameters and T: GFLOPS) of baseline and pruning Siamese networks on ReID datasets.}
\scalebox{1}{%
\begin{tabular}{|l|c|c||r|r||r|r||r|r|}
\hline
& & & \multicolumn{2}{c||}{\textbf{Market-1501}} & \multicolumn{2}{c||}{\textbf{DukeMTMC}}& \multicolumn{2}{c|}{\textbf{CUHK03-NP}} \\ \cline{4-9} 
\textbf{Networks}& \textbf{M} & \textbf{T} & \textbf{mAP } & \textbf{rank-1 }  & \textbf{mAP } & \textbf{rank-1 }  & \textbf{mAP } & \textbf{rank-1}   \\ \hline \hline
ResNet50 & 23.48 & 6.32& 69.16 & 85.07 & 59.46 & 76.39 & 47.57 & 48.43      \\\hline
ResNet34 & 21.28 & 6.67 & 67.44 & 84.09  & 58.36 & 75.45 & 45.51 & 47.14\\\hline
ResNet18 & 11.12 & 3.09 & 61.23 & 81.18 & 52.07 & 71.63  & 38.27 & 39.57\\\hline \hline
L1 & 11.90 & 2.96 & 67.04 & 84.71  & 57.51 & 75.00   & 44.08 & 46.50  \\\hline
Taylor  & 12.09 & 3.21 & 66.35 & 84.44  & 57.90 & 75.72 &  44.40 & 46.21\\\hline
AutoBalanced & 11.90 & 2.96& 65.46 & 83.64 & 56.45 & 74.64 & 41.85 & 44.21\\\hline
Entropy  & 11.90 & 2.96 & 65.16 & 82.39  & 56.64 & 74.64 & 42.44 & 44.07\\\hline
PSFP & 11.90 & 2.96 &65.92 & 83.72 & 56.96 & 74.66 & 42.38 & 45.58\\\hline

\end{tabular}%
}

\label{Market1501 Triplet Results}
\end{table*}

\begin{figure*}[t!]
\centering\includegraphics[height=1\linewidth]{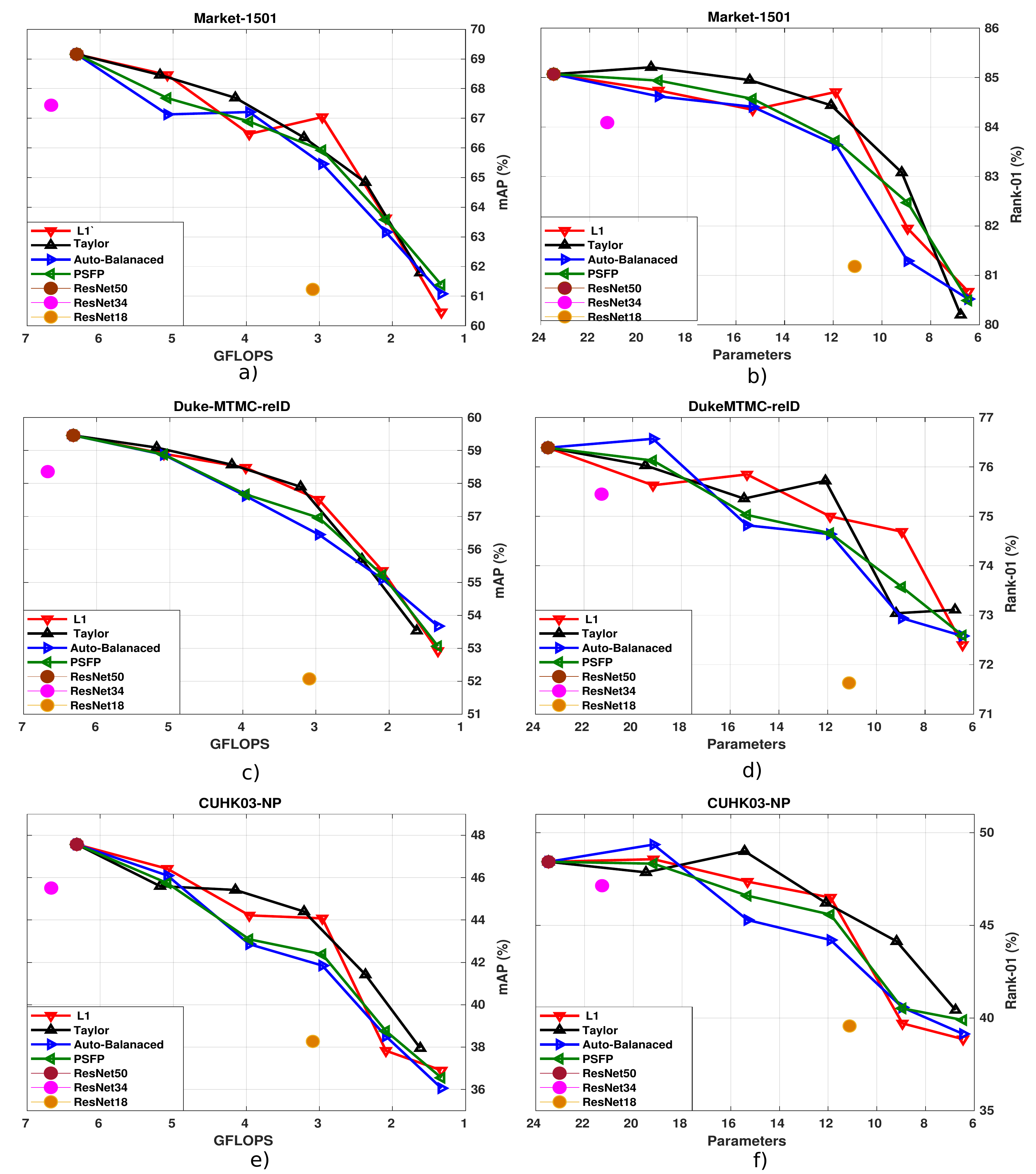}
\caption{Comparative ReID performance analysis of the pruning methods for all the ReID datasets: (a,c,e) mAP vs GFLOPS and (b,d,f) Parameters vs Rank-01.}
\label{fig:prun}
\end{figure*}

The pruned models also have shown less performance drop in terms of accuracy while reducing considerably the number of FLOPS and parameters. Pruned models are faster than backbone ResNet50 network while having similar performance (around 1\%). Plus, the pruned models have a similar number of FLOPS and parameters to ResNet18 while having better results on the three performance metrics. This means that pruning a larger model is more advantageous than using a shallower model like ResNet18.

To get a more global view of these results,  the graphics in Figures~\ref{fig:prun} depicts visually which models are better where the optimal placement would be top right and the worse would be bottom left. There are two graphics for each dataset where the first one presents the mAP vs FLOPS and the second one presents Rank1 vs Parameters.

\subsection{Pruning on target application data with weak ReID baseline :}

The objective of this experiment is to analyze and compare the pruning techniques with weak ReID baseline such as \textit{Trinet}~\citep{hermans2017defense}. Table~\ref{Scenarios_all} reports the experimental evaluations of all the scenarios. For fair comparison, we chose to keep the compression ratio to 5\% of the total number of channels. Our \texttt{Scenario 2} results are produced using the HaoLi Iteration 3 model as the model pruned on the pretraining dataset. Using the same model for the considered techniques gives us a better idea on which pruning technique is the best when we pruned directly on our task dataset.
\begin{table*}[t!]
\centering
\caption{ Comparison of network accuracy (mAP and  rank-1) and complexity (memory: M (parameters) and time: T (GFLOPS)) with different pruning Scenarios on all the person ReID datasets.}
\begin{adjustbox}{width=1\textwidth}
\begin{tabular}{|l|c||c|c|c|c||c|c|r|r||r|r|r|r|r|}
\hline
& & \multicolumn{4}{c||}{\textbf{Market-1501}} & \multicolumn{4}{c||}{\textbf{DukeMTMC-reID}}& \multicolumn{4}{c|}{\textbf{CUHK03-NP}}  \\ \cline{3-14} 
\textbf{ Methods}& \textbf{ \small{Scenario}} & \textbf{\small{mAP }} & \textbf{\small{rank-1 } } &  \textbf{\small{T}} & \textbf{\small{M}}  & \textbf{\small{mAP }} & \textbf{\small{rank-1 } }&  \textbf{\small{T}} & \textbf{\small{M}}  &  \textbf{\small{mAP }} &\textbf{\small{rank-1 }} &  \textbf{\small{T}} & \textbf{\small{M}}   \\ \hline \hline
L1        &      & 67.04  & 84.71   & 2.96  &11.90   & 57.51  &75.00  & 2.96   &11.90  &44.08  &46.50    & 2.96  & 11.90  \\
Taylor    &     & 66.35  & 84.44   & 3.21  & 12.09 & 57.90   &75.72  & 3.21   &12.09 &44.40   &46.21   & 3.21  &12.09\\
Entropy      &  1    & 65.16  & 82.39   & 2.96  & 11.90  & 56.64  &74.64  & 2.96   &11.90  &42.44  &44.07   & 2.96  &11.90\\
Auto-Balanced&      & 65.46  & 83.64   & 2.96  & 11.90  & 56.45  &74.64  & 2.96   & 11.90 &41.85  &44.21   & 2.96  & 11.90\\
PSFP         &      &65.92   & 83.72   & 2.96  & 11.90  & 56.96  &74.66  & 2.96   & 11.90 &42.38  &45.58   & 2.96  & 11.90\\ \hline 

L1        &      & 49.18  & 70.67   & 2.09  & 8.95  & 40.91  &61.67  & 2.09   & 8.95 &26.29   &27.36   & 2.09 & 8.95 \\
Taylor    &      & 32.03  & 53.92   & 2.11  & 8.31  & 23.54  &40.44  & 2.02   &7.99  &13.61   &13.14   & 2.05 & 8.09\\
Entropy      &    2  & 7.38   & 17.31   & 2.09  & 8.95  & 2.34   &6.78   & 2.09   &8.95  &7.83    &7.29    & 2.09 & 8.95\\
Auto-Balanced&      &47.36   & 68.74   & 2.09  & 8.95  & 43.19  &64.14  & 2.09   &8.95  &26.69   &27.36   & 2.09 & 8.95\\
PSFP         &      & 65.03  & 80.85   & 2.09  & 8.95  & 53.05  &73.20   & 2.09   & 8.95 &42.19   &43.86   & 2.09 & 8.95 \\ \hline  

L1        &      & 67.44  & 84.23   & 5.08  & 19.17 & 57.16  &74.28  & 5.08   & 19.17 &44.73  &47.86   & 5.08 & 19.17\\
Taylor    &      & 63.29  & 81.38   & 5.28  & 19.60  & 44.14  &63.33  & 4.98   & 18.53 &36.91  &38.86   & 4.96 & 18.46\\
Entropy      &    3  & 60.27  & 79.99   & 5.08  & 19.17 & 52.62  &71.72  & 5.08   & 19.17 & 38.5  &40.14   & 5.08 & 19.17\\
Auto-Balanced&      & 67.49  & 84.26   & 5.08  & 19.17 & 58.14  &75.27  & 5.08   & 19.17 &46.54  &48.07   & 5.08 & 19.17\\
PSFP         &      & 67.68  & 84.78   & 5.08  & 19.17 & 57.51  &74.87  & 5.08   & 19.17 &46.25  &48.20    & 5.08 & 19.17\\ \hline  

L1        &      & 31.41  & 55.70    & 5.08  & 19.17 & 27.94  &48.79  & 5.08   & 19.17 &14.97  &16.07   & 5.08 & 19.17\\
Taylor    &      & 6.39   & 12.89   & 5.14  & 19.10  & 1.18   &2.29   & 4.80   & 17.89 & 6.28  &5.71    & 4.93 & 18.35\\
Entropy      &    4  & 25.41  & 46.35   & 5.08  & 19.17 &21.08   &41.16  & 5.08   & 19.17 &11.31  &11.64   & 5.08 & 19.17\\
Auto-Balanced&      & 59.27  & 78.80    & 5.08  & 19.17 &49.64   &67.77  & 5.08   & 19.17 &36.67  &38.79   & 5.08 & 19.17\\
PSFP         &      & N/A    & N/A     & N/A   & N/A   &N/A     &N/A    & N/A    &N/A    & N/A   &N/A     & N/A  &  N/A \\ \hline
 
\end{tabular}%

\end{adjustbox}
\label{Scenarios_all}
\end{table*}

As we can observe in Table~\ref{Scenarios_all}, the results with pruning directly on the target operational domain are not performing as good as the the performances of the same pruned model with large-scale pre-training dataset. We can make following observations from these results: (1) pruning and fine-tuning should be done on the same domain as in the case of \texttt{ Scenario 1} and \texttt{Scenario 3}, no matter whether it is source or target operational domain; (2) lack of data in target domain effect the pruning accuracy  to regain the information loss by the pruning of the weak channels; (3) with the large-scale source dataset and the \texttt{L1} method, we were able to prune our model to the same number of FLOPS as \texttt{Scenario 2} (2.09 GFLOPS) but our Rank1 accuracy was 81.95\% instead of 70.67\%. The \texttt{L1} method also seems to be better suited to prune directly on the task dataset compared to \texttt{Taylor} and \texttt{Entropy}. This might  be explained by the fact we don't have many samples per person since \texttt{Taylor} and \texttt{Entropy} approach uses a subset of samples to determine which channels to prune compared to the \texttt{L1} method that ranks the channels with their weights; (4) \texttt{Scenario 4} is not viable since all methods performance drop drastically. (5) as for the \texttt{Auto-Balanced} and the \texttt{PSFP techniques}, they seem to outperform the other methods. This could be explained by the fact that auto-balanced modifies the loss in order to transfer the information of the pruned channels to the remaining one. This scheme seems to help considerably when our number of samples is limited. The \texttt{PSFP} seems to be the best suited algorithm to pruned models on a limited dataset. This can probably be explained by the fact that we only zeroized the pruned channels which keeps the model architecture which allows the recovery of certain soft-pruned channels during the fine-tuning phrase. Our results with the \texttt{PSFP} also very similar to the ones obtain with the \texttt{Scenario 1} scheme where we prune our models on the large-scale source dataset and then fine-tune on our task domain dataset. The great advantage of this method is the fact we can prune and fine-tune our models in the same step. Plus, we\textquoteright re skipping the slow step of pruning on the very large ImageNet dataset.

To compare the scenarios further, we used two compression ratios which are around half the FLOPS (C1) and around one third (C2) of the FLOPS of the original ResNet50 model. The \texttt{Scenario 2} model for the first compression is using the second iteration \texttt{L1} model as the model pruned on ImageNet. As for the second compression, We\textquoteright re using the third iteration. The results for the following experiments are found in Tables \ref{Scenarios_Compare_Market1501}.

\begin{table*}[t!]
\centering
\caption{Comparison of network accuracy (mAP and  rank-1) and complexity (M: memory (parameters)  and time: T (GFLOPS)) with different pruning compression ratios of different scenarios on all the ReID datasets.}
\label{Scenarios_Compare_Market1501}
\begin{adjustbox}{width=1\textwidth}
\begin{tabular}{|c|c|c|c|c|c|c|c|c|c|c|}
\hline
\multicolumn{1}{|c|}{\textbf{Datasets}} & \multicolumn{1}{|c|}{\textbf{Scenario}} & \multicolumn{1}{|c|}{\textbf{M}} & \multicolumn{1}{|c|}{\textbf{T}} & \multicolumn{2}{c|}{\textbf{L1}}       & \multicolumn{2}{c|}{\textbf{Auto-Balanced}} & \multicolumn{2}{c|}{\textbf{PSFP}}        \\ \cline{5-10} 
 & (compression)   &  &                                 & \textbf{mAP } & \textbf{rank-1} & \textbf{mAP }  & \textbf{rank-1}  & \textbf{mAP} & \textbf{rank-1} \\ \hline \hline
 &1 (C1)   &11.90 & 2.96                                         & 67.04 & 84.71                              & 65.46 & 83.64                              &65.92  & 83.72                               \\
 &2 (C1) &11.90 & 2.96                                           &47.46 & 69.36                              & 47.57& 70.46                                & 65.55 & 82.69                               \\
\textbf{Market-1501} &3 (C1)  &11.90& 2.96                                         &53.22 & 72.21                               & 54.73 & 74.05                                & 65.88 & 82.19                               \\ \cline{2-10}
 &1 (C2)  &8.95& 2.09                                        &63.63 & 81.95                               &  63.16 & 81.29                                & 63.58 & 82.47                               \\
 &2 (C2) &8.95& 2.09                                         &49.18 & 70.67                               & 47.36 & 68.74                                & 65.03 & 80.85                               \\
  &3 (C2) & 8.95& 2.09                                        &41.93 & 62.62                              & 48.30 & 69.00                                    & 65.88 & 82.91                               \\ \hline

  &1 (C1)                 &11.90                      & 2.96  &57.51   & 75.00                            &56.64   & 74.64                    &56.96 & 74.66                               \\
 &2 (C1)                    &11.90                    & 2.96   &40.60   & 59.47                           &41.09    & 61.09                             &53.71 & 71.90                                \\
  \textbf{Duke-MTMC} &3 (C1)                 &11.90                         & 2.96  & 46.88  & 65.93                           &45.54   & 66.16                              &56.62 & 74.09                               \\ \cline{2-10}
 &1 (C2)                & 8.95                        & 2.09 &55.35  & 74.69                           &55.10   & 72.94                               &55.22 & 73.57                               \\
 &2 (C2)                & 8.95                        & 2.09 & 40.91     & 61.67                          & 43.19     & 64.14                                & 53.05 & 73.20                               \\
 &3 (C2)              & 8.95                           & 2.09  & 39.17  & 58.71                         & 34.80     & 54.58                                 &56.77& 73.38                               \\ \hline
 
 &1 (C1)                             & 11.90           & 2.96   & 44.08   & 46.50                   & 41.85            & 44.21                          & 42.38        & 45.58                               \\
  &2 (C1)                             & 11.90          & 2.96    & 27.23 & 28.43                &    27.44           & 29.57                             & 40.47     & 45.00                                   \\
 \textbf{CUHK03-NP} &3 (C1)                                  & 11.90    & 2.96  & 33.57    & 36.07                &    33.34            & 35.29                              & 40.66  & 44.57                                \\   \cline{2-10}
 &1 (C2)                                & 8.95         & 2.09 & 37.83 & 39.71                &      38.51          & 40.57                                & 38.76  & 40.52                               \\
 &2 (C2)                                 & 8.95     & 2.09    & 26.29 & 27.36                &     26.69           & 27.36                                  & 42.19 & 43.86                                \\
 &3 (C2)                                & 8.95     & 2.09    & 26.79 & 28.29                &      26.50          & 27.14                                 & 40.31 & 40.14                               \\ \hline

\end{tabular}%
\end{adjustbox}

\end{table*}

Tables \ref{Scenarios_Compare_Market1501} shows us that \texttt{Scenario 1} is truly the best one since all the results outperform the other ones for any method and any dataset. As for the comparison between the \texttt{Scenario 2 and 3}, the conclusion to determine which one is better is hard to make since \texttt{ Scenario 2} can be done using many configurations to get to a model similar to the one in \texttt{Scenario 3}. We could either prune more on the large-scale source dataset or prune less. The \texttt{Scenario 2} results are also affected by the choice of the pruned model on the large-scale source set. Our first compression results using the second iteration of the L1 method performs less in terms of recognition accuracy than pruning only on the target operational dataset (\texttt{Scenario 3}). But using the third iteration  as shown in the second compression results, our \texttt{Scenario 2} results are better than our \texttt{Scenario 3} results.\\

\subsection{Pruning on target application data with strong ReID baseline :}

The results shown in Table \ref{Scenarios_Compare_Market1501} indicate that PSFP so far is the best performing pruning approach in most scenarios. Additionally, since PSFP is suitable for deploying compressed model -- training can be done while the pruning is applied -- we  apply this technique on a strong ReID baseline. Therefore, the aim of this experiment is to analyze the effectiveness of the pruning techniques using a strong ReID baseline  PCB~\citep{sun2018beyond}. We show two experimental analysis with PSFP pruning of PCB architectures. The first experiment shows the effect of ReID accuracy while pruning only backbone feature extractor (indicated as PCB (BFE)), while the second one considers all the layers (i.e. local convolutional layers and Fully connected layers) after backbone architectures those perhaps use for feature compression and for classifications tasks. In addition to the original ResNet~\cite{he2016deep}, we also performed experiments with SE-ResNet~\cite{hu2018squeeze} to see the effectiveness of pruning methods on different backbone CNNs.

Experimental results for the strong baseline PCB are reported in Table~\ref{strong_baseline} both for Market-1501 and DukeMTMC-reID datasets. Our results shows a consistency with the initial claims --  the number of FLOPS and parameters required by PCB's ResNet and SE-ResNet architectures are reduced by half, while maintaining a comparable rank-1 accuracy for both ReID datasets. Results also suggest that PSFP pruning of local convolutional layers and FC layers have little effect on ReID accuracy as the margin of differences between \textit{PSFP+PCB(BFE)} and  \textit{PSFP+PCB(BFE+LC+FC)} is small. This analysis implies that it is worth pruning backbone architecture rather with local convolutional and FC layers since it allows more memory and parameters reduction. It is worth noting that the margin of decline in mAP accuracy is higher than that of rank-1 accuracy for both backbones, and on all ReID datasets.\\

\begin{table}[t!]
\centering
\caption{ Comparison of network accuracy (mAP and  rank-1) and complexity (M: Parameters and T: GFLOPS) with different pruning Scenarios on all the person ReID datasets. BFE: Backbone Feature Extractor; LC : Local Convloutional Layer; FC: Fully Connected Layer. }
\begin{adjustbox}{width=1.0\textwidth}
 \begin{tabular}{|l|l|c|c|c|c||c|c|c|r|r|r|r|r|}
\hline
 & &\multicolumn{4}{c||}{\textbf{Market-1501}} & \multicolumn{4}{c||}{\textbf{DukeMTMC-reID}} \\ \cline{3-10} 
\textbf{ Methods} & Backbone &\textbf{\small{mAP }} & \textbf{\small{rank-1 } } &  \textbf{\small{T}} & \textbf{\small{M}}  & \textbf{\small{mAP }} & \textbf{\small{rank-1 } }&  \textbf{\small{T}} & \textbf{\small{M}}     \\ \hline \hline
PCB (Baseline)    &  & 77.3  & 92.4  & 6.1  & 27.2   & 65.3 & 81.9  & 6.1   & 27.2    \\
PSFP + PCB (BFE)  &  ResNet50   & 69.4  & 90.4   & 2.6  & 12.0   & 60.7 & 78.5 & 2.6   & 12.0    \\ 
PSFP + PCB (BFE+ LC + FC)  &    & 69.1  & 89.1   & 2.6  & 11.16   & 59.4 & 77.8 & 2.6   & 11.16     \\ \hline

PCB (Baseline)    & &  77.3 & 91.7  & 5.9  & 27.7   & 65.3 & 81.2  & 5.9   & 27.7    \\

PSFP + PCB (BFE)  &   SE-ResNet50  & 70.0  & 88.4   & 2.59  & 14.5   & 61.9 & 78.9  & 2.59   & 14.5  \\

PSFP + PCB (BFE+ LC + FC) & & 69.2   &  87.2 & 2.58   & 13.69 & 59.4  & 77.4  &  2.58 & 13.69     \\ \hline
\end{tabular}%
\end{adjustbox}
\label{strong_baseline}
\end{table}

\begin{table}[t!]
\centering
\caption{ Comparison of network accuracy (mAP and  rank-1) and complexity (M: Parameters and T: GFLOPS) with different pruning criteria of PSFP approach  on all the person ReID datasets. BFE: Backbone Feature Extractor.}
\begin{adjustbox}{width=1.0\textwidth}
 \begin{tabular}{|l|l||c|c|c|c||c|c|c|r|r|r|r|r|}
\hline
 & &\multicolumn{4}{c||}{\textbf{Market-1501}} & \multicolumn{4}{c||}{\textbf{DukeMTMC-reID}} \\ \cline{3-10} 
\textbf{ Methods} & Criteria &\textbf{\small{mAP }} & \textbf{\small{rank-1 } } &  \textbf{\small{T}} & \textbf{\small{M}}  & \textbf{\small{mAP }} & \textbf{\small{rank-1 } }&  \textbf{\small{T}} & \textbf{\small{M}}     \\ \hline \hline

PSFP+PCB(BFE)  &  $l_1$-norm  & 68.7  & 90.0   & 2.6  & 12.0   & 60.6 & 78.8  & 2.6   & 12.0  \\
\hline

PSFP+PCB(BFE)  & $l_2$-norm   & 69.4  & 90.4   & 2.6  & 12.0   & 60.7 & 78.5 & 2.6   & 12.0    \\
\hline
 \end{tabular}%
\end{adjustbox}
\label{strong_baseline_1}
\end{table}

\textbf{Filter selection criteria.}
As a part of ablation study, this experiment aims to analyze the effect of magnitude based filter selection criteria such as $l_p$-norm on ReID accuracy. We conduct this experiment with PSFP+PCB(BFE) on ReNet50. We show a comparative ReID performance analysis between $l_1$-norm and $l_2$-norm on Table~\ref{strong_baseline_1}. It can be observed from  Table~\ref{strong_baseline_1} that the ReID performance of $l_2$-norm criteria is marginally better than that of $l_1$-norm criteria. This is due to the effect of the largest element that has been dominant in $l_2$-norm. As a consequence, the filters with largest weights preserved while pruning provide more discriminative feature for better recognition accuracy.

\begin{figure*}[t!]
\centering\includegraphics[width=1\linewidth]{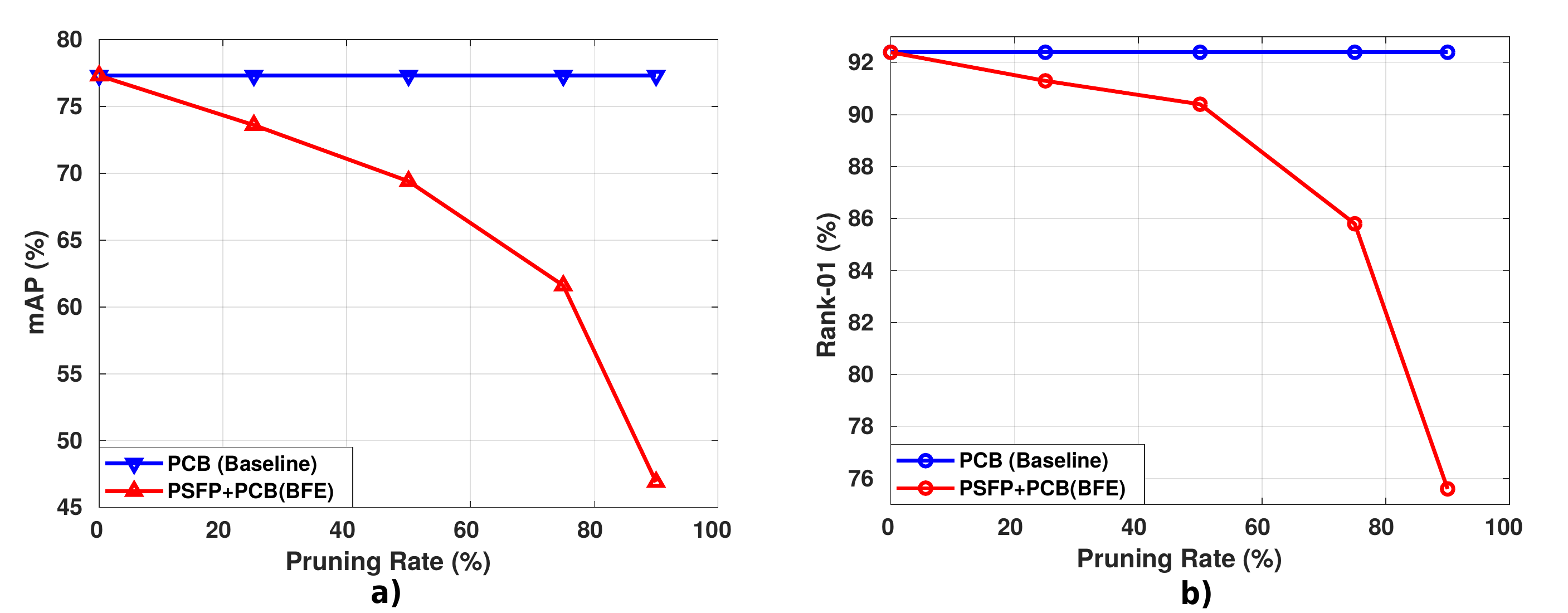}
\caption{Comparative ReID performance analysis of the         PSFP+PCB(BFE) methods with different pruning rates for Market-1501 dataset: (a) mAP vs  pruning rates and (b) Rank-01 vs Pruning rates.}
\label{fig:prun_rate}
\end{figure*}

\textbf{Varying pruning rates.}
The objective of this experiment is to observe ReID performance when varying pruning rates. It was performed with PSFP+PCB(BFE). Figures~\ref{fig:prun_rate}(a) and (b) show the mAP and rank-01 accuracy obtained with varying pruning rates, respectively. With both measures, the accuracy of the pruned model drops exponentially with growing pruning rates. For pruning rates between 0\% and 25\%, the accuracy of the pruned model drops marginally. The pruning rate above 50\% leads to drastic decline in ReID performance. When pruning a larger number of filters, the loss of information affects accuracy considerably.

To further analysis the pruning on target operational domain,  we apply the best fine-tuning practices proposed in \citep{Finetune_Best}, as presented in Section \ref{S:Prune_CNN_Design}. we\textquoteright ve calculated their metrics for ImageNet and Market11501 and got $0.005$ for the cosine distance and $2.45$ for MMD. With these metrics and the fact that Market1501 has fewer than 20 samples for each class, the authors proposed to freeze the feature extractor during the fine-tuning to avoid over fitting since our task dataset is small and close to our large-scale pretraining dataset. Since our problem of a small dataset was showing during the retraining phase of the pruned network, we decided to try prune one layer at a time and freeze the others during the retraining phase. The goal of this strategy is to force the pruned layers to relearn the loss information while maintaining the other layers in the same optimal region as the baseline model. This method was tried for \texttt{Scenario 2} with the \texttt{L1} method. We decided to prune the layer 5 of the ResNet50 while freezing the rest of the network. The model was pruned to 2.61 GFLOPS and the rank1 accuracy was 76.10\%. This experiment shows that we could limit the effects of pruning by using a layer by layer approach and freezing the other layers to regain the accuracy. The problem with this scheme is that it is not very effective time-wise since it\textquoteright s a long and fastidious task to prune and retrain to the desired compression ratio for each layer instead of doing the whole model in one pass.

\section{Conclusion}
\label{S:5}
 In this paper, we exploit the prunability of the state-of-the-art pruning models that are suitable for compressing deep architecture for person ReID application in terms of criteria to select channels, and of strategies to reduce channels. In addition to that, we propose different scenarios or pipelines for leveraging a pruning method during the deployment of a network for a target application. Experimental evaluations on multiple benchmarks source and target datasets indicate that pruning can considerably reduce network complexity (number of FLOPS and parameters) while maintaining a high level of accuracy. It also suggests that pruning larger CNNs can also provide a significantly better performance than fine-tuning smaller ones. One key observation of the scenario-based experimental evaluations is that pruning and fine-tuning should be performed in the same domain.  
 
 Future experiments could explore a reduction in pruning iterations in order to reduced the impact of pruning on knowledge corruption. Retraining of the pruned networks could also be improved by adding a learning rate decay. Using layer-by-layer methods, with different compression ratios for each layer can improve the results since some layers are more resilient to pruning than others. Techniques for freezing parts of the network can also improve accuracy, but drastically increase the time complexity for  pruning and retraining phases.  The soft pruning method could also benefit from better selection criteria, e.g., using a gradient-based approach instead of the norm of the channel weights. Finally, another interesting future experiment would be to avoid costly pruning on large pre-training dataset, and only use the progressive soft pruning scheme to see if can achieve similar results with higher compression ratios.
While pruning have proven to be effective in person ReID, we realized that it is focused on a same domain, which can limit its usage. In addition, work on pruning in the unsupervised learning settings is still quite limited. Future work could extended such pruning methods to unsupervised domain adaptation in person ReID.

\begin{backmatter}

\bibliographystyle{bmc-mathphys} 
\bibliography{bmc_article}      

\section*{Availability of data and materials}
The re-identification datasets used to validate the findings of this work are publicly available that can be be downloaded following the given references of each dataset.

\section*{Competing interests}
The authors declare that they have no competing interests.

\section*{Funding}
This work was supported by the Mitacs Accelerate Master's Fellowship, Elevate Postdoctoral Fellowship Program, and the Natural Sciences and Engineering Research Council of Canada.

\section*{Authors' contributions}
All the authors contributed by participating in the discussion, experimentation and drafting the manuscript of the work described in this paper. All authors read and approved the final manuscript.

\section*{Acknowledgements}
The authors would like to thank all the state-of-the-art works for sharing their code which help us to validate our approach by setting their approaches as baselines.

\end{backmatter}
\end{document}